\documentclass[12pt,letterpaper]{article}

\usepackage{amsmath}
\usepackage{amssymb}
\usepackage{amsfonts}
\usepackage{graphicx}
\usepackage{algorithm}
\usepackage{algorithmic}
\usepackage{url}
\usepackage{comment}
\usepackage[toc,page]{appendix}
\usepackage{authblk}
\usepackage{setspace}
\usepackage{color}

\newcommand{\no}{\nonumber} 
\newcommand{\bb}{\boldsymbol}

\def\mysingleq#1{`#1'}

\title{Multiple-view clustering for identifying subject clusters and brain sub-networks using functional connectivity matrices without vectorization}
\author[* 1, 4]{Tomoki Tokuda }
\author[1, 3]{Okito Yamashita }
\author[2, 1]{Junichiro Yoshimoto }
\affil[1]{\small Brain Information Communication Research Laboratory Group, Advanced Telecommunications Research Institutes International, 2-2-2 Hikaridai, Seika-cho, Soraku-gun, 
Kyoto 619-0288, Japan}
\affil[2]{\small Nara Institute of Science and Technology, 8916-5 Takayama, Ikoma, Nara 630-0192, Japan}
\affil[3]{\small Center for Advanced Intelligence Project, RIKEN, Nihonbashi 1-chome Mitsui Building, 15th floor, 1-4-1 Nihonbashi, Chuo-ku, Tokyo 103-0027, Japan}
\affil[4]{\small Okinawa Institute of Science and Technology Graduate University, 1919-1 Tancha, 
Okinawa 904-0495, Japan}
\affil[*]{\small Corresponding author, t-tokuda@atr.jp }

\date{\vspace{-5ex}}

\providecommand{\keywords}[1]{\textbf{\textit{Keywords---}} #1}

\begin{document}

\maketitle

\begin{abstract}
In neuroscience, the functional magnetic resonance imaging (fMRI) is a vital tool to non-invasively access brain activity. Using fMRI, the functional connectivity (FC) between brain regions can be inferred, which has contributed to a number of findings of the fundamental properties of the brain. As an important clinical application of FC, clustering of subjects based on FC recently draws much attention, which can potentially reveal important heterogeneity in subjects such as subtypes of psychiatric disorders. In particular, a multiple-view clustering method is a powerful analytical tool, which identifies clustering patterns of subjects depending on their FC in specific brain areas. However, when one applies an existing multiple-view clustering method to fMRI data, there is a need to simplify the data structure, independently dealing with elements in a FC matrix, i.e., vectorizing a correlation matrix. Such a simplification may distort the clustering results. To overcome this problem, we propose a novel multiple-view clustering method based on Wishart mixture models, which preserves the correlation matrix structure without vectorization. The uniqueness of this method is that the multiple-view clustering of subjects is based on particular networks of nodes (or regions of interest, ROIs), optimized in a data-driven manner. Hence, it can identify multiple underlying pairs of associations between a subject cluster solution and a ROI sub-network. The key assumption of the method is independence among sub-networks, which is effectively addressed by whitening correlation matrices. We applied the proposed method to synthetic and fMRI data, demonstrating the usefulness and power of the proposed method.
\end{abstract}
\vspace{5mm}

\section{Introduction}
In neuroscience, the functional magnetic resonance imaging (fMRI) is a vital tool to non-invasively access brain activity. In particular, the functional connectivity (FC) between different brain regions, which is inferred from fMRI, is quite useful to investigate fundamental properties of the brain network, and contributed to the  uncovering of intriguing associations between neural circuits and their functions
\cite{smitha2017resting, tie2014defining, agosta2012resting, finn2015functional, filippi2019resting}. 
FC is also promising as potential biomarkers for various psychiatric disorders \cite{ichikawa2020primary, yamashita2020generalizable}. Furthermore, it is recently suggested that the clustering of patients based on FC may reveal important subtypes of psychiatric disorders \cite{drysdale2017resting, clementz2016identification, saad2020review, miranda2020functional}. 
However, despite its usefulness, 
high dimensionality of FC data hinders an effective application of conventional clustering methods. This is mainly because in addition to the well-known problem of \mysingleq{curse of dimensionality} \cite{mwangi2014review} for high-dimensional datasets, there are specific problems for cluster analysis. Typically, there is a complex structure of data with multiple subject clustering patterns, depending on the selection of features \cite{tzortzis2010multiple, yang2018multi}.
That is, there may exist the underlying multiple cluster solutions of subjects, which are characterized by different combinations of features. 

To overcome this difficulty, a multiple-view clustering method has been proposed, which is successfully applied to fMRI data to identify subtypes of depressive disorder \cite{tokuda2017multiple, tokuda2018identification}. This method's main principle is to optimally partition features into several groups (called \mysingleq{view}) in which subject clustering is performed separately. In so doing, the partitioning of features and subject clustering are performed simultaneously. Hence, the yielded results are optimal in terms of both feature partition and subject clustering. The method is based on a Gaussian mixture model in which a univariate Gaussian distribution is fitted to each cluster block.

Despite its usefulness and wide areas of application, there is a drawback of the method when it is applied to FC data from fMRI.
Typically, FC denotes a Pearson's correlation of blood-oxygenation-level-dependent (BOLD) signals between ROIs, which is represented in the form of a matrix. 
The relationship among elements of the correlation matrix need to satisfy the positive definite condition, which restricts the existence of a correlation matrix in non-Euclidean space \cite{pennec2006riemannian} (mathematically, such space is not closed under the operation of subtraction; the resulting matrix may not be positive definite anymore \cite{johnson1970positive}). Hence, extracting elements from a correlation matrix and separately dealing with these elements in vector form may distort the underlying cluster structures. Nonetheless, element-wise feature extraction from a correlation matrix is a conventional practice for the clustering due to computational simplicity and high interpretability.

In the present paper,  we propose a novel multiple-view clustering method for correlation matrices, which keeps the correlation structure intact, hence not resulting in the distortion mentioned above. 
Here, we use the terminology of \mysingleq{node} in a graph theory as a synonym for ROI. For a mathematical formulation, it is also a synonym of \mysingleq{variable}, which denotes a generative model of a correlation matrix.
In our approach, nodes in a correlation matrix are exclusively partitioned into several views, each of which enables the identification of a sub-network in a graph that is represented by a given correlation matrix. Further, we assume an object clustering structure for each network fitted by mixture models of Wishart distribution. The number of views as well as the number of object clusters are automatically estimated in the nonparametric Bayesian framework.
Such a node-based approach allows us to extend a mixture model from a single view to multiple views in a natural way while preserving the positive definite structure of correlation matrices. The proposed method provides a novel perspective of clustering in terms of network analysis in a data-driven manner. 

In the following sections, we first review related works on our proposed method. Next, we present our novel multiple-view clustering method. Subsequently, we evaluate the performance of the proposed method by applying it to artificial data. Further, we apply the proposed method to fMRI data, which evaluates the proposed method's performance for task fMRI data and resting-state fMRI data. Finally, we discuss interpretations of the yielded results from a neuroscientific point of view and important features of the proposed method. 

\section{Related work}
There are several approaches to revealing the underlying multiple-view structure in data (for comprehensive reviews see \cite{bailey2013alternative, hu2017finding}). We can largely categorize these approaches into \mysingleq{alternative clustering} and \mysingleq{subspace clustering}. In the alternative clustering, different cluster solutions from a given cluster solution are sequentially identified \cite{bae2006coala, jain2008simultaneous, dang2010generation, dang2015framework}. The idea behind it is to search the alternative cluster solution that is orthogonal to the given cluster solution. 
Typically, the alternative clustering is inferred based on the dissimilarity from the given cluster solution, using all features. On the other hand, in the subspace clustering, clustering is performed in several subspaces of features. This approach further entails two major strategies for identification of the subspace: the de-coupled strategy and the coupled strategy \cite{hu2018subspace}. For the de-coupled strategy, the appropriate subspace is identified prior to performing clustering \cite{cui2007non, qi2009principled, ye2016generalized, wang2019multiple}, whereas for the coupled strategy the identification of the subspace and clustering are performed simultaneously. In the latter strategy, there are several variants in terms of clustering methods. In \cite{niu2010multiple}, the multiple cluster solutions are inferred by spectral clustering based on optimization of the non-redundant subspace. 
In \cite{mautz2018discovering}, the rotation of the feature space and the projection coupled with K-means clustering are simultaneously optimized. For the model-based approach, the Gaussian mixture model is extended to multiple-view structure \cite{guan2010variational, tokuda2017multiple}. The proposed method in the present paper belongs to this category of multiple-view clustering.
Despite such large variations in multiple-view clustering, to the best of our knowledge, there has been no attempt to perform multiple-view clustering for correlation matrices while keeping the correlation matrix structure intact. 

Next, we briefly review the recent progress on the Wishart mixture model, which is a basic probabilistic model for our proposed method. There have been major theoretical developments with regard to the Wishart mixture model despite limited attempts to apply to real data. The expectation-maximization algorithm has been derived for inferring relevant parameters \cite{hidot2010expectation, saint2013new}, whereas a Bayesian nonparametric extension for inferring the number of clusters is also formulated \cite{cherian2015bayesian}. However, the area of application has been limited to computer vision analysis \cite{hidot2010expectation, ferro2001unsupervised}. Recently, in the field of neuroscience, a Wishart mixture model was applied to develop a hidden Markov model in the context of dynamic FC \cite{nielsen2016nonparametric, nielsen2018predictive}. 

In a nutshell, our proposed method is theoretically quite unique, revealing the underlying multiple-view structure based on the Wishart mixture model. Practically, it is also a first attempt to apply a multiple-view clustering method to FC correlation matrix without vectorization.

\begin{table}[!]
\centering
\caption{Notation for multiple-view clustering method}
\begin{tabular}{|c|c|l|}
\hline
Domain & Notation & Description \\ 
\cline{1-3}
Data & $n$ & Sample size \\
& $p$ & Number of nodes \\
& $ \bb{M}$& $p \times p$ covariance matrix \\
& $ \bb{M}_v$& Covariance matrix for nodes in view $v$\\
& $ \bb{M}_{v, g}$& Covariance matrix for nodes in node cluster $g$ in view $v$\\
& $ \bb{R}$& $p \times p$ correlation matrix \\
& $ \bb{D}$& $n$ collection of $\bb{M}$ (or $\bb{R}$) \\
\cline{1-3}
Model & $V$ & Number of views \\
&$K_v$& Number of object clusters in view $v$ \\
&$G_v$& Number of node clusters in view $v$ \\
& $\bb{u}$ & $p$-dimensional vector of view memberships \\
& $\bb{z}_v$ & $n$-dimensional vector of object cluster memberships \\
& & in view $v$ \\
& $\bb{y}_v$ & $p_v$-dimensional vector of node cluster memberships \\
& & in view $v$ where $p_v$ is the number of nodes in view $v$ \\
& $\bb{\Sigma}_{v, k}$& Scale parameter for object cluster $k$ in view $v$ \\
& $\bb{\Sigma}_{v, k, g}$& Scale parameter for object cluster $k$ and \\
&& node cluster $g$ in view $v$\\
& $T$ & Degree of freedom\\ 
\hline
\end{tabular}
\label{parameters}
\end{table}

\section{Method}
\subsection{What is multiple-view clustering?}
Multiple-view clustering is based on the idea that different cluster solutions of objects (subjects) exist depending on the features that we focus on. In a toy example in Fig.\ref{Illustration}a, we have different cluster solutions for six letters, depending on the color, style, and number of holes, which corresponds to views 1-3, respectively. We apply this idea to the clustering of objects characterized by correlation matrices (Fig.\ref{Illustration}b). There are two underlying object cluster solutions in this illustration, depending on two subsets of nodes (nodes 1 and 2). In this example, the nodes are sorted in nodes 1 and 2, which facilitates visual inspection of the object cluster structure. If we focus on patterns of correlations in nodes 1, we have a cluster solution, as in view 1. On the other hand, if we focus on patterns in nodes 2, we have another cluster solution, as in view 2. An important assumption in this model is that the underlying correlation of nodes between nodes 1 and 2 is zero. This suggests that we consider the clustering of objects for each independent sub-network of nodes (Fig.\ref{Illustration}c).

In real data, of course, it is not a trivial question of how to partition nodes. In the present study, assuming an exclusive partition of nodes where each node belongs to only a single view, we aim to simultaneously optimize a partition of nodes and object cluster solutions using a stochastic model.

\subsection{Model}
The proposed multiple-view clustering method is based on a probability model, which assumes that covariance matrices follow a Wishart distribution conditional on a cluster block. We consider that the Wishart distribution is the most natural and direct probabilistic formulation of FC correlation matrix as in the same spirit as \cite{nielsen2018predictive}.
In this section, we outline the model, which explicitly incorporates node partition and object clustering (for parameters in the model, please refer to Table~\ref{parameters}).

We consider a $p\times T$ dimensional data matrix, typically, time series data, denoted as $p$-dimensional $\bb{X}(t)$ at time point $t$ ($t=1, \ldots, T$).
Here, we suppose $p$ nodes, each of which generates a single time series data with length $T$. In the context of fMRI data, one may suppose BOLD signals where the number of ROIs is $p$. We assume that $\bb{X}(t)$ is generated from a multivariate Gaussian distribution,
\begin{eqnarray}
\bb{X}(t) & \overset{i.i.d.}{\sim} & N(\bb{\mu}, \bb{\Sigma}),
\label{timeseries}
\end{eqnarray}
where $N(\bb{\mu}, \bb{\Sigma})$ denotes a multivariate Gaussian distribution with mean vector $\bb{\mu}$ and covariance matrix $\bb{\Sigma}$.
Further, assuming $\bb{\mu}=\bb{0}$, the $p \times p$ matrix $\bb{M} = \sum_{t=1}^T \bb{X}(t)\bb{X}(t)'$ follows a Wishart distribution, which is given by
\begin{eqnarray}
\mathcal{W}(\bb{M} | T, \bb{\Sigma}) = C_{\bb{M}, T} \times g(\bb{M}, T, \bb{\Sigma}),
\label{decompo}
\end{eqnarray}
where $T$ and $ \bb{\Sigma}$ are the degree of freedom and a scale parameter, respectively. Here, the constant term $C_{\bb{M}, T}$ and the non-constant term $g(\cdot)$ are defined as
\begin{eqnarray}
\no C_{\bb{M}, T} &=& \frac{|\bb{M}|^{(T-p-1)/2}}{2^{pT/2} \Gamma_p(T/2)}\\
\no g(\bb{M}, T, \bb{\Sigma}) &=&\frac{\exp (-\mbox{tr}(\bb{\Sigma} ^{-1}\bb{M})/2)}{|\bb{\Sigma} |^{T/2}},
\end{eqnarray}
where $\Gamma_p(a) =\pi^{p(p-1)/4}\prod_{j=1}^{p}\Gamma((a-j+1)/2)$ with a gamma function $\Gamma(\cdot)$.

To model multiple-view clustering, we introduce several structures in the Wishart model in Eq.(\ref{decompo}). First, we introduce a view structure for a scale parameter $\bb{\Sigma}$ in Eq.(\ref{decompo}). We partition nodes into several groups (referred to as 'view' hereafter), assuming that the time series between views are independent. To explicitly denote a partition of nodes, we consider a $p$-dimensional vector of view partition $\bb{u}$. This index defines view memberships of nodes, denoting that node $i$ belongs to view $u(i)$. For instance, $\bb{u}=(1, 1, 1, 2, 2)$ denotes that nodes one, two, and three belong to view one, whereas nodes four and five view two. Because of the assumption of independence between views, the scale parameter $\bb{\Sigma}$ in Eq.(\ref{decompo}) is partitioned into $V$ views ($v=1, \ldots, V$) as follows:
\begin{eqnarray}
\bb{\Sigma} =
\begin{pmatrix}
\bb{\Sigma}_{1} & \ldots & \bb{0} \\
\ldots & \bb{\Sigma}_{v} & \ldots \\
\bb{0} & \ldots & \bb{\Sigma}_{V} \\
\end{pmatrix}
,
\label{matrixstr}
\end{eqnarray}
where $\bb{\Sigma}_{v}$ is a scale parameter among the nodes that belong to view $v$. Note that for simplicity, nodes are sorted in ascending order of view memberships in Eq.(\ref{matrixstr}). Conditional on the view structure $\bb{u}$, it can be shown that the Wishart density function is given by
\begin{eqnarray}
f(\bb{M}|T, \bb{\Sigma}, \bb{u}) &=& C_{\bb{M}, T} \prod_{v=1}^V g(\bb{M}_v, T, \bb{\Sigma}_v),
\label{multiden}
\end{eqnarray}
where $\bb{M}_v$ is a covariance matrix for nodes in view $v$. It is worth noting that $C_{\bb{M}, T}$ does not depend on the view structure $\bb{u}$, which facilitates the inference of relevant parameters.

Second, we consider a collection of $n$ covariance matrices, denoted as $\bb{D} = \{\bb{M}^{(i)}\}_{i=1}^n$. In the context of fMRI, one may suppose an FC matrix for $n$ subjects. Assuming that the distribution of a covariance matrix follows a Wishart distribution, we cluster objects by fitting each realization of $\bb{M}$ to a finite mixture of Wishart distribution with $K$ components,
\begin{eqnarray}
f(\bb{M}|\bb{\theta}) = \sum_{k=1}^K w_k \times \mathcal{W}(\bb{M}|T_k, \bb{\Sigma}_{*, k}),
\label{mixmodel}
\end{eqnarray}
where $T_k$ and $\bb{\Sigma}_{*, k}$ are the degree of freedom and a scale parameter for component $k$, and $w_k$ is the mixing proportion ($\sum_{k=1}^K w_k=1$). Note that the suffix $*$ in $\bb{\Sigma}_{*, k}$ denotes that the scale parameter is for all nodes without partitioning. For simplicity, we set $T_k=T$ ($k=1, \ldots, K$). Note that we have not considered a view structure yet. The parameter $\bb{\theta}$ on the left-hand side in Eq.(\ref{mixmodel}) denotes all relevant parameters for the Wishart mixture model on the right-hand side. At this stage, the number of components $K$ is fixed. Next, we introduce a binary $K$-dimensional vector $\bb{z}$, which denotes a latent class label of an object, that is, $z(k) = 1$ if the object belongs to component $k$, and $z(k)= 0$ otherwise. Conditional on the latent class label $\bb{z}$, the Wishart density function is given by
\begin{eqnarray}
\no f(\bb{M}|\bb{\theta}, \bb{z}) &=& \prod_{k=1}^K \mathcal{W}(\bb{M}| T, \bb{\Sigma}_{*, k})^{z(k)} \\
&=&C_{\bb{M}, T} \prod_{k=1}^K g(\bb{M}, T, \bb{\Sigma}_{*, k})^{z(k)}.
\label{mixWishartcondi}
\end{eqnarray}
This equation indicates that the density function is identical to the Wishart density function in the component $k$ such that $z(k) = 1$.

Next, we incorporate the view structure into Eq.({\ref{mixWishartcondi}).
Analog to Eq.(\ref{multiden}), the density function conditional on the view-dependent object cluster membership $\bb{z}_v$ becomes
\begin{eqnarray}
f(\bb{M}|\bb{\theta}, \bb{u}, \{\bb{z}_v\}) &=& C_{\bb{M}, T} \prod_{v=1}^V
\prod_{k=1}^{K_v} g(\bb{M}_v, T, \bb{\Sigma}_{v, k})^{z_v(k)},
\label{multiden2}
\end{eqnarray}
where $K_v$ is the number of object clusters in view $v$.

Moreover, in the spirit of co-clustering \cite{tokuda2017multiple}, we also introduce a cluster structure for nodes within a view. This has the effect of modeling more than one independent network within a view, hence identifying all relevant networks for the object clustering in the same view. Here, to avoid cluttering, we do not lay it out, but the following model development is basically the same (see Appendix \ref{nodeclustering} for details).

Finally, using Eq.(\ref{multiden2}), the likelihood of data $\bb{D}$ conditional on view memberships and object cluster memberships is given by
\begin{eqnarray}
\no && f(\bb{D}|T, \{ \bb{\Sigma}_{v, k} \}, \bb{u}, \{\bb{z}_v\}) \\
&&~~~~~~~~~~~= \prod_{i=1}^n C_{\bb{M}^{(i)}, T} \prod_{v=1}^V \prod_{k=1}^{K_v}
g(\bb{M}_{v}^{(i)}, T, \bb{\Sigma}_{v, k})^{z_v(k)}.
\label{likelihood}
\end{eqnarray}
This equation indicates that the density function of each covariance matrix is evaluated by the Wishart distribution with appropriate scale parameters, which are defined by view membership and subject cluster membership. It is noted that Eq.(\ref{likelihood}) is in a simple form of multiplication because of the assumption of independence among views in Eq.(\ref{matrixstr}). Also, it is of note that we consider the degree of freedom $T$ as a parameter to estimate. In usual circumstances, the degree of freedom $T$ is known in advance. However, if the independent assumption of $\bb{X}(t)$ in Eq.(\ref{timeseries}) does not hold, we need to consider the effective degree of freedom, which is often the case for fMRI time series data \cite{fox2005human, afyouni2019effective}. In the present method, we allow for flexibility of this quantity, which is inferred in the framework of model estimation.

\subsection{Correlation matrix}
We consider the application of the aforementioned model to correlation matrices. We assume that a correlation matrix $\bb{R}$ is given as $\bb{R} = \sum_{t=1}^T\bb{X}(t)\bb{X}(t)'/T$, where $p$-dimensional vector $\bb{X}(t)$ is generated as in Eq.(\ref{timeseries}), with the diagonal elements in covariance matrix $\bb{\Sigma}$ being one. It can be interpreted that the correlation matrix $\bb{R}$ is a normalized covariance matrix. The effect of such normalization is appropriately considered in prior distributions for scale parameters in the following section. Further, it is noted that the diagonal elements of $\bb{R}$ are close to one, but not exactly one in general because of the stochastic nature of $\bb{X}(t)$. More specifically, denoting the $v$th element of $\bb{X}(t)$ as $x_{v}(t)$, the diagonal element $(v, v)$ of $\bb{R}$ is given by $\sum_{t=1}^T x_{v}(t)^2/T$. From the Gaussian assumption of $\bb{X}(t)$, the numerator of this quantity follows a $\chi^2$ distribution with degree of freedom $T$ (hence, the expectation of $\sum_{t=1}^T x_{v}(t)^2/T$ is one).
In contrast, the diagonal elements of an empirical correlation matrix are one (constant). The exact modeling of such an empirical correlation matrix is not straightforward, which involves a further complication of the model \cite{zhang2006sampling}. Hence, as an approximate model, we apply the model mentioned above in the previous section to an empirical correlation matrix.

\subsection{Priors}
For Bayesian inference, we introduce prior distributions for relevant parameters.
First, for a scale parameter $\bb{\Sigma}_{v, k}$, denoting the matrix size as $p' \times p'$, we assume a conjugate prior, that is, an inverse Wishart distribution $\mathcal{W}^{-1}(\cdot|\cdot)$,
\begin{eqnarray}
\no \bb{\Sigma}_{v, k} \sim \mathcal{W}^{-1}(\cdot | \bb{S}, \nu),
\end{eqnarray}
where $\nu$ and $\bb{S}$ are the degree of freedom and scale parameter, respectively. We used a non-informative setting $\nu=p'+3$ for the degree of freedom. When we apply the model to the correlation matrices, we set the scale parameter $\bb{S}=(\nu-p'-1)\bb{I}/T$, where $\bb{I}$ is an identity matrix.
With this setting of the scale parameter $\bb{S}$, the expectation of diagonal elements of $\bb{\Sigma}_{v, k}$ becomes $\bb{I}/T$. Hence, the prior expectation of an observed matrix becomes an identity matrix, which fits well with a correlation matrix.

Second, for both view and cluster labels $\bb{u}$ and $\bb{z}$, we assume a Chinese restaurant process (CRP), which enables the estimation of the number of views and clusters:
\begin{eqnarray}
\no \bb{u} &\sim& \mbox{CRP}(\alpha)\\
\no \bb{z}_v &\sim& \mbox{CRP}(\alpha),
\end{eqnarray}
where $\bb{z}_v$ denotes the subject cluster memberships for view $v$. We set the hyperparameter $\alpha$ (concentration) in CRP to one as a default (see Appendix \ref{appenCRP} for more details).

Third, for the degree of freedom $T$, we assume a (uniform) categorical distribution $\mbox{Cat} (\cdot | \cdot)$,
\begin{eqnarray}
\no T \sim \mbox{Cat}(\cdot|\bb{c}, \bb{p}),
\end{eqnarray}
where $\bb{c}$ and $\bb{p}$ are the sample space and event probability, respectively. We set $\bb{c} = (T_1, \ldots, T_q)$ and that $p_i = 1/q$ ($i=1, \ldots, q$) (see Appendix \ref{degree} for more details).

Lastly, we assume that all these priors are independent. Hence, the joint prior distribution is simply given by the products of these priors.

\subsection{Inference}
By Bayes' rule, the posterior distribution in our model is given by
\begin{eqnarray}
\no f(T, \{ \bb{\Sigma}_{v, k} \}, \bb{u}, \{\bb{z}_v\} |
\bb{D}) &\sim&
f(\bb{D}|T, \{ \bb{\Sigma}_{v, k} \}, \bb{u}, \{\bb{z}_v\}), \\
&&
\no ~~~~~~\times f(T, \{ \bb{\Sigma}_{v, k} \}, \bb{u}, \{\bb{z}_v\}),
\end{eqnarray}
where the first term on the right-hand side is given by Eq.(\ref{likelihood}). The second is the joint prior for the relevant parameters.
For efficient optimization of the posterior, we integrate out all scale parameters $\{ \bb{\Sigma}_{v, k} \}$:
\begin{eqnarray}
\no L &\equiv& f(T, \bb{u}, \{\bb{z}_v\} |
\bb{D}) \\
&&~~~~~~~= \int f(T, \{ \bb{\Sigma}_{v, k} \}, \bb{u}, \{\bb{z}_v\} | \bb{D}) d \{ \bb{\Sigma}_{v, k} \}.
\label{lossfunction}
\end{eqnarray}
Because we assume a conjugate prior for scale parameters $\{ \bb{\Sigma}_{v, k} \}$, this integration can be obtained in closed form. Hence, our objective is to find the remainder of the parameters $T$, $\bb{v}$, and $\{\bb{z}_v\}$ that optimize (maximize) $L$ in Eq.(\ref{lossfunction}).
There are several approaches to optimizing $L$, such as variational Bayes, Gibbs sampling, and iterative conditional modes (ICM) \cite{blei2006variational, little2019machine}. In the present study, we optimize $L$ by means of ICM, which in turn gives maximum posterior (MAP) estimates of $T$, $\bb{u}$, and $\{\bb{z}_v\}$. This algorithm iteratively optimizes parameters one by one until $L$ converges. Because there is no guarantee for global optimization in the algorithm, we repeat this procedure for a number of initializations (setting the number of initializations $J$ to 1000 as a default). Among the parameter estimates for various initializations, we select the one that gives the maximum value of $L$ (see Appendix \ref{appenalgo} for details).

The motivation for taking this approach is that such a hard assignment of view membership allows for a substantial reduction in computation time because of the explicit independent structure of the scale parameters in Eq.(\ref{matrixstr}) (for the computing complexity see section \ref{discussion2}). Averaging methods such as variational Bays would deal with view memberships as \mysingleq{averaging}. This is not computationally efficient in our context because there is a need to evaluate the whole matrix with high dimensions. It is also noted that Gibbs sampling is not efficient, although it gives a hard assignment of view memberships because it takes a long time to evaluate the convergence of chains. Rather than simulating in long chains, we make the best use of parallel computing, which allows for an efficient search in MAP utilizing ICM.

\subsection{Preprocessing}
To effectively perform our proposed method, we undertook the following preprocessing of the data. One is the regularization of covariance matrices in a high-dimensional case (large $p$). The other is the whitening of covariance matrices, which transforms covariance matrices to meet the assumption of independence between views. Both preprocesses apply to the correlation matrices as well.

\subsubsection{Regularization}
For a data matrix $\bb{X}$ in Eq.(\ref{timeseries}), an empirical covariance matrix becomes singular when the number of data points is smaller than the number of nodes. In such a case, we cannot evaluate the covariance matrix using a Wishart distribution. Even if the number of data points is sufficiently large, the same ill-defined case may occur when the independent assumption in terms of datapoint $t$ does not hold, which is typically the case for fMRI data. Hence, in a high-dimensional case (a large number of nodes), some regularization of the covariance matrices is required. In the present study, we use a regularization method proposed by \cite{ledoit2004well}, which effectively shrinks the sample eigenvalues.

\subsubsection{Whitening}
The proposed method largely relies on the assumption that nodes are independent between views. This assumption may not hold in real data analysis, in particular, in fMRI data. Nonetheless, even in that case, there may be different object cluster structures between views. To reveal such structures using the proposed method, we consider the whitening of covariance matrices \cite{varoquaux2010detection, dadi2019benchmarking}.
The whitening procedure entails the following linear transformation of a covariance matrix:
\begin{eqnarray}
\widetilde{\bb{M}}^{(i)} \leftarrow \bar{\bb{M}}^{-1/2} \bb{M}^{(i)} \bar{\bb{M}}^{-1/2}
\label{white2}
\end{eqnarray}
where $\bar{\bb{M}}$ is the empirical mean covariance matrix of data, i.e.,
$\bar{\bb{M}}=\sum_{i=1}^n \bb{M}^{(i)} /n$; $\bar{\bb{M}}^{-1/2} = \bb{U}' \bb{\Delta}^{-1/2} \bb{U}$, where $\bb{U}$ and $\bb{\Delta}$ are given by eigenvalue decomposition $\bb{U}' \bb{\Delta} \bb{U} = \bar{\bb{M}}$. Here, the prime denotes the transposition of a matrix. Through the whitening procedure, it is expected that if there is no object cluster structure in off-diagonal elements in the covariance matrix, these elements become (nearly) zero for all objects. In contrast, if there is an object cluster structure, these elements do not become zero for all objects. Hence, whitening has the effect of making elements zeros between different views. Incorporation of the whitening procedure into the proposed method considerably widens the area of application to real data. Note that in the case of correlation matrix, we further transform the yielded matrix $\widetilde{\bb{M}}^{(i)} $ into the correlation matrix.

Regarding whitening, one may wonder whether there is a correspondence between the whitened nodes and the original nodes. In other words, does node $j$ of $\widetilde{\bb{M}}^{(i)}$ represent the node $j$ of $\bb{M}^{(i)}$ in Eq.(\ref{white2})?
The whitening procedure in question is the one that maximizes the sum of the cross-correlation between the whitened and original nodes \cite{kessy2018optimal}.
It is taken for granted that a correspondence exists between whitened nodes and original nodes in the literature, for example, in terms of the affine framework \cite{ng2014transport}. However, as whitening involves different scaling for nodes, 
the order of data points is not, in general, preserved between the whitened and original nodes. 
To the best of our knowledge, there is no theoretical foundation to ensure complete agreement between two node spaces in terms of cross-correlations.
Hence, for interpretations of whitened ROIs, we consider the following two approaches. One approach is to consider that whitened ROIs represent the corresponding original ROIs (hereafter, referred to as the \mysingleq{heuristic} approach). One may examine cross-correlations between original ROIs and whitened ROIs, which are given by $\bar{\bb{M}}^{1/2}$ \cite{kessy2018optimal}. The other approach is to consider the effect of whitening (hereafter, referred to as the \mysingleq{vigorous} approach) as follows. For a given original ROI $i$, we evaluate its contribution to a particular view (e.g., view $v$)
as the summation of absolute values of correlation coefficients between the ROI in question and the whitened ROIs in the view. Denoting such a contribution as Importance ($i$, $v$), we formulate it as
\begin{eqnarray}
\mbox{Importance}~(i, v) = \sum_{u~\in~\mbox{view $v$}} |r_{i, u}|,
\label{imp}
\end{eqnarray}
where $r_{i, u}$ is the cross-correlation between the original ROI $i$ and whitened ROI $u$. Importance can be used to select relevant ROIs for this view.

\section{Results}
In this section, we report the results of two simulation studies to evaluate the proposed method's performance. One is the application to synthetic data, whereas the other is to fMRI data. For synthetic data, we aim to examine the proposed method's workability and compare its performance with other methods. For fMRI data, we aim to demonstrate the workability of the proposed method in a real situation. To this end, we consider the application to two types of datasets released in the Human Connectome Project (HCP) \cite{wu20171200}. The first datasets are task fMRI data. We mixed subjects into two different tasks. Hence, two clusters are artificially embedded in the data. We tested whether the proposed method can recover these \mysingleq{true} clusters characterized by the difference in tasks. Second, we applied the method to resting-state fMRI data. We considered two sessions of resting-state fMRI of the same subjects. We examined whether the proposed method can identify similar subject clusters between the two datasets. In both cases, we evaluated a correlation matrix for a subject based on the mean BOLD signal in brain parcels of Shen parcellation \cite{shen2013groupwise}, which defines 268 ROIs in the whole brain.

\subsection{Synthetic data}
\subsubsection {Data generation}
We consider two types of data: Type 1 and Type 2. For Type 1, we assume that the background correlation is zero, which exactly matches our model assumption. On the other hand, for Type 2, we assume that the background correlation is not zero, which does not match the model assumption and hence requires a whitening procedure. For these two types of data, we used the following settings of the view and cluster structures:
\begin{itemize}
\item Number of nodes: $p=30$
\item Number of objects: $n=100$
\item Number of views: $V=3 $ (each view has ten nodes)
\item Number of object clusters: $K_v=4 ~ (v=1, \ldots, V)$ (hence, each cluster has 25 objects)
\item Number of node clusters: $G_v=1 ~ (v=1, \ldots, V)$
\end{itemize}

We generate data based on Eq.(\ref{timeseries}), using a correlation matrix
$\bb{\Sigma}^*$, which is obtained by combining the correlation matrices in three views. For each object cluster in a view, a $10 \times 10$ correlation matrix is generated as follows:
First, a $10 \times 10$ lower triangular matrix $\bb{L}$ is generated, where each element is randomly generated from the standard Gaussian distribution (mean 0 and variance 1). Second, we evaluate the matrix product $\bb{A}=\bb{L}*\bb{L}'$. Because of the property of Cholesky decomposition, the yielded matrix $\bb{A}$ is positive definite. Third, we transform this covariance matrix into a correlation matrix. Subsequently, we randomly shuffle the nodes in the correlation matrix. We generate such a correlation matrix for each object cluster in each view. Finally, we make a correlation matrix for all nodes by diagonally combining the correlation matrices from each view, setting correlations of nodes between views to zero as in Eq.(\ref{matrixstr}). Such a combination is specific for object cluster-ID and view ID. Denoting the yielded matrix as $\bb{\Sigma}$ (we omit a symbol for dependency on object cluster structures), we obtain $\bb{\Sigma}^*$ by adding a \mysingleq{noisy} background matrix $\bb{N}$ (diagonal elements are one),
\begin{eqnarray}
\bb{\Sigma}^* = (1-w) \times \bb{\Sigma} + w \times \bb{N},
\label{noiseratio}
\end{eqnarray}
where weight $w$ is the noise ratio that we manipulate. Regarding the background matrix $\bb{N}$, we consider two cases for its off-diagonal elements: 0 (Type 1) and 0.2 (Type 2). Finally, we set the number of data points to $T = p+10$.

Next, for each object, we randomly assign a cluster membership of each view.
Subsequently, using the corresponding correlation matrix $\bb{\Sigma}^*$ for cluster memberships, the data (sample size $T$) are randomly generated for each object from a multivariate Gaussian distribution in Eq.(\ref{timeseries}). We obtain a collection of correlation matrices from the generated data, which are used as inputs for clustering methods. We have 100 replications for each configuration. As a preprocessing step, we do not perform a whitening procedure for datasets of Type 1, whereas for datasets of Type 2, we perform a whitening procedure.

\subsubsection{Other clustering methods}
To the best of our knowledge, no method exists that simultaneously
infers both view and object cluster structures for correlation matrices. Hence, as competitive methods, we consider several combinations of existing clustering methods.

\subsubsection*{Hierarchical + K-means}
In this method, a hierarchical clustering method is first performed for partitioning nodes using the mean correlation matrix over objects as in the sub-network analysis of 
ROIs \cite{salvador2005neurophysiological, lee2013resting}.
Subsequently, K-means clustering is performed for the clustering of objects using vectorized correlation matrices in each view. For hierarchical clustering, we set the link function to \mysingleq{average} with
the distance defined as $(1-r)$, where $r$ is a corresponding off-diagonal element of the correlation matrix.

\subsubsection*{Community detection + K-means}
Instead of a hierarchical clustering method, a community detection method \cite{aicher2015learning} is performed for the partitioning of nodes as in the sub-network analysis of \cite{lynn2019physics}, followed by the clustering of objects using a K-means clustering method. The community detection method is based on the weighted stochastic block model (WSBM), a generalized stochastic block model for weighted edges.

\subsubsection*{K-means}
We consider that the input is vectorized correlation matrices. 
Without considering a view structure, we naively apply a K-means clustering method to vectorized correlation matrices. 

\subsubsection*{Multiple-view clustering based on Gaussian mixture models}
This is another type of multiple-view clustering, where the input is vectorized correlation matrices as in the case of \mysingleq{K-means} above.  Instead of nodes, the matrix elements are partitioned into views. Specifically, we use the multiple-view clustering method based on Gaussian mixture models \cite{tokuda2017multiple}, which fits Gaussian mixture models to the matrix elements in each view.


\vspace{\baselineskip}

\noindent For the hierarchical + K-means, and community detection + K-means methods, we fixed the number of views to the true one. In addition, for all competitive methods that use K-means, we fixed the number of object clusters to the true one.
On the other hand, for our proposed method and multiple-view clustering method based on Gaussian mixtures, these quantities were not fixed, but were inferred in a data-driven manner. When K-means and multiple-view clustering based on Gaussian mixture models were applied, features were standardized (i.e., mean zero and standard deviation one), following the conventional procedure. To alleviate the local minimum problem of the object function, 
we set the number of initializations to 100 for all clustering methods in the present study.

\subsubsection{Evaluation of performance}
We evaluated the performance of these clustering methods in two ways. One is the recovery of the true view structure, whereas the other is the recovery of the true object cluster structures.
To recover the true view structure, we examined the agreement between the true view memberships and estimated view memberships. We evaluated the mean value of the adjusted rand index (ARI) \cite{hubert1985comparing} of view memberships over 100 replications of synthetic data. To evaluate the recovery of the true object cluster structures, we examined the extent to which the true object clusters in each view were recovered. Because view labels are arbitrary, that is, they are simply nominal, there is generally no correspondence between the numbering of the true view labels and the numbering of yielded view labels. Hence, we need to match these views. However, such a matching is not necessarily obvious, particularly when the clustering performance exacerbates. For each true view, we simply evaluated the recovery of the true object cluster structure from the perspective of taking the maximum value of ARI with estimated cluster solutions in all views. Lastly, we took the grand average of such ARIs over different true views and over 100 replications to be the overall performance.

\subsubsection{Results of performance}
The performance results are summarized in Fig.\ref{simres}. For Type 1 data (i.e., background correlation zero between views), the proposed method (\mysingleq{Multiple} in Fig.\ref{simres}) performed reasonably well, perfectly recovering both the view and object cluster structures for the range of noise weight $w$ between 0 and 0.6. However, the performance deteriorates as the noise weight $w$ further increases. The method of community detection + K-means (\mysingleq{WSBM-Kmeans}) performs better than the proposed method for the noise weight of $w$ between 0.7 and 0.9. In contrast, the hierarchical + K-means (\mysingleq{Hier-Kmeans}) and K-means (\mysingleq{K-means}) methods do not perform well. In particular, the multiple-view clustering method based on Gaussian mixture models yielded poor results.

For Type 2 data (i.e., background correlation 0.2 between views), the performance of the proposed method is as good as in the case of Type 1 data. However, the performance of the method of community detection + K-means worsens. This is because the true view structure becomes vague in the mean correlation matrix for Type 2 data, which hinders this method from correctly identifying the true view and object cluster structures. As in the case of Type 1 data, the multiple-view clustering method based on Gaussian mixture models performed poorly.

Overall, this simulation study suggests the reasonably good performance of the proposed method. Importantly, its performance is stable regardless of whether the background correlation is zero or non-zero. On the other hand, the competitive methods do not perform well, particularly in Type 2 data. The main reason for poor performance is the lack of information on the view structure in the mean correlation matrix. This suggests that without explicitly modeling view structures, there is a limit to recovering the view structure from the data. Lastly, it is of note that the multiple-view clustering method based on Gaussian mixture models performs rather poorly. This implies that for this method the multiple-view structure largely disappears through vectorization and standardization of the given correlation matrices.


\subsection{Task fMRI}
In this experiment, we applied the proposed method to samples of correlation matrices obtained from task fMRI data in the HCP data depository. The objective is to examine whether the proposed method can identify subject clusters characterized by task differences. That is, we aim to identify the relevant view(s) for the task differences. 

In HCP data, the task fMRI data consists of seven tasks: motor, working memory, language, social cognition, gambling, relational processing, and emotion processing. Each subject is engaged in all these tasks. Among these seven tasks, we focused on three tasks: motor (MOTOR), working memory (WM), and language processing (LANGUAGE), which have a relatively longer duration of scanning than the other tasks. Because our objective was to examine the performance of our proposed method, we used a simple structure of a dataset in which subjects differed only in two tasks. Hence, there were two clusters in a dataset. The data generation procedure was as follows. For each task, we randomly selected 50 subjects from the HCP depository (970 subjects) without overlapping, which resulted in three task-specific datasets. We combined two of these three datasets, which results in three mixed datasets: MOTOR-WM, MOTOR-LANGUAGE, and LANGUAGE-WM, each of which consisted of 100 subjects. We applied the proposed method to these three types of mixed datasets.

It is of note that this way of randomly selecting subjects affects controlling factors other than tasks. Alternatively, one might choose 50 subjects and extract two task data for each subject, which results in a sample size of 100. This procedure would perfectly control factors other than the task difference. However, the samples would not be independent because we would have used two data sets for the same subject. To avoid such dependency between samples, we opted to select 100 different subjects in the manner mentioned above, which is analog to an actual data collection situation.

To evaluate correlation matrices, we used the minimally preprocessed task fMRI data provided by the HCP (e.g., \path{tfMRI_MOTOR_LR.nii}). Unlike resting-state fMRI, it would be appropriate to remove the influence of task-related activities by regressing out of task events. Nonetheless, it has been reported that such a regression step has only minimal effects on FC analysis \cite{cole2014intrinsic}. Moreover, in the present study, our objective was to examine the feasibility of the clustering method. Hence, for simplicity, we used the whole time series, including the time duration of non-tasks. We removed linear components related to 12 motion parameters, low-pass filtering, and scrubbing as part of preprocessing.

Regarding the proposed method, we performed regularization and whitening for the correlation matrices. The number of initializations $J$ was set to 1000. To select a stable model, we used the following heuristic for model selection: First, we selected the top five models in terms of the posterior $L$ in Eq.(\ref{lossfunction}). Among these five models, we subsequently evaluated the agreement of view memberships between models by means of ARI. Then, we identified the pair of models that give the largest value of ARI. The final model was the one in this pair, which gave a larger value of the posterior $L$. This procedure allowed us to choose a stable model with a large value of the posterior, removing the possibility of a spurious model.

\subsubsection{Results of clustering}
Here, we report on the clustering results in detail for the MOTOR-WM dataset. This performed the best in terms of recovery of the task difference. First, the number of views was estimated as 25, whereas the number of ROIs in the views ranged from 1 to 32 (Fig.\ref{summary1}). The number of subject clusters ranged from 1 to 56 (Fig.\ref{summary1}, \ref{summary2}). Second, we examined the stability of clustering results in terms of view memberships. To this end, we focused on 30 best models out of 1000 random initializations. We evaluated the stability of view memberships between the optimal model and those 30 models by means of the Dice coefficient \cite{sorensen1948method} (Fig.\ref{stability}; the red points denote $95 \%$ upper limits by permutation test). It was observed that the extent of stability varied depending on the view. However, it appeared to be more stable as the number of subject clusters increased.

\subsubsection{Recovery of task difference}
In this experiment, the dataset consisted of two heterogeneous groups of subjects attributed to different tasks, MOTOR, and WM. We examined whether such a heterogeneous structure of subjects was recovered in the clustering results, which was the main objective of this experiment. It was found that in view 16, the proposed method best recovered the task difference (Fig.\ref{recovery}). In Fig.\ref{stability}, it can also be observed that view 16 is among the stable views. Furthermore, to confirm the recovery of the task difference,
we examined the distributions of subjects in view 16 (Fig.\ref{subjectcluster}). Remarkably, subject cluster 1 was dominated by subjects with the task of WM, whereas subject cluster 2 was dominated by those with the task of MOTOR. Additionally, to examine whether such a clear task difference can be recovered without the multiple-view structure, we applied the proposed method by imposing on the constraint of the single view structure (i.e., the number of views is not estimated, but fixed to one). It is found that the clear task difference was not recovered in this case.

\subsubsection{Relevant brain regions for view 16 in MOTOR-WM}
Next, we analyzed relevant brain regions for view 16. This view consisted of 12 (whitened) ROIs (ROI ID: 21, 23, 24, 25, 26, 27, 33, 158, 159, 167, 171, and 174 in Shen parcellation). For further analysis, we focused on subject clusters 1 and 2, which are dominated by subjects with the WM and MOTOR tasks, respectively. Visualization of the mean correlation matrices for these ROIs shows that the correlation matrix patterns differ between subject clusters 1 and 2 (Fig.\ref{white}). Differences, though subtle, are also observed in several instances of the subjects as well (columns 2 to 5 in Fig.\ref{white}).

Lastly, we identified the relevant sub-network of the brain in this view. We considered both heuristic and vigorous approaches to the selection of ROIs. We selected the aforementioned 12 ROIs in the heuristic approach, confirming that the original ROIs and whitened ROIs were in good agreement (Fig.\ref{correspandencewhite}). In addition, in the vigorous approach, the top 12 ROIs in terms of the importance index in Eq.(\ref{imp}) is the same as the 12 ROIs identified in the heuristic approach (Fig.\ref{rig}). A large discrepancy between the 12th ROI and the 13th ROI in Fig.\ref{rig} suggests that the main contribution of the original ROI space is due to these top 12 ROIs. Regarding location in the brain, it was found that these ROIs are related to the motor strip and parietal regions (Fig.\ref{brain}). More detail about these brain regions is discussed in section \ref{interpretation}.

\subsection{Resting-state fMRI}
In this experiment, we applied the proposed method to resting-state fMRI data. We aimed to examine the reproducibility of the clustering results between two resting-state fMRI data obtained from the same subjects. Moreover, inspired by the clustering results (shown in the following section), connectome fingerprinting has been examined \cite{pannunzi2017resting}. The study by \cite{pannunzi2017resting} demonstrated that a subject was identifiable using a specific collection of FCs in predefined brain networks. In contrast, we approached connectome fingerprinting in terms of the clustering of subjects without prior information on the brain network structures.

In the HCP data, resting-state fMRI was acquired four times for a single subject: twice for each phase encoding L-R and R-L \cite{wu20171200}. In the present experiment, we used two L-R phase encoding data (\path{REST1_LR_hp2000_clean.nii} and \path{REST2_LR_hp2000_clean.nii}) to prepare two datasets of correlation matrices for 100 randomly selected subjects. We refer to these datasets as \mysingleq{Rest1} and \mysingleq{Rest2}, respectively. We performed the same preprocessing as in the case of task fMRI datasets in the previous section, followed by applying the proposed method to each dataset separately. It is of note that for the whitening of correlation matrices, we used the mean correlation matrix of all subjects in both Rest1 and Rest2. This allows us to compare the clustering results between the two datasets because the whitened ROI space becomes the same for Rest1 and Rest2.

\subsubsection{Results of clustering}
The number of views is estimated as 24 and 29 for Rest1 and Rest2, respectively
(Fig.\ref{summary1rest}). For Rest1, the number of ROIs in a view ranged from 2 to 29, while for Rest2, it ranged from 4 to 45. There is a large variation over views regarding the number of subject clusters; it ranges from 2 to 52 for Rest1 and 2 to 50 for Rest2.
(Fig.\ref{summary1rest}, Fig.\ref{summary2rest}).

Next, we examined the clustering results' stability as in the case of task fMRI, focusing on the 30 best models for each dataset (Fig.\ref{stabilityrest}). We evaluated the stability of view memberships between the optimal model and the 30 models using the Dice coefficient. The extent of stability varied depending on the view. However, there was a general tendency to be more stable as the number of subject clusters increased.

\subsubsection{Comparison between Rest1 and Rest2}
We identified pairs of views of Rest1 and Rest2 in terms of both agreements of view memberships and subject cluster memberships between Rest1 and Rest2 (Fig.\ref{comparisonrest}). It was found that the agreement in the following pairs of views was significant based on permutation tests ($p<0.05$) in both cases:
\begin{itemize}
\item Pair 1: View 24 in Rest1 and view 27 in Rest2
\item Pair 2: View 23 in Rest1 and view 29 in Rest2
\item Pair 3: View 22 in Rest1 and view 28 in Rest2
\end{itemize}

For these pairs, there was good agreement between the subject cluster memberships between Rest1 and Rest2, whereas the number of subject clusters in a view was considerably large.
Hence, it was hypothesized that there might be good agreement at the level of individual subjects. To examine this hypothesis, we evaluated the accuracy of the matching of subjects between Rest1 and Rest2, which is analogous to the functional connectome fingerprint \cite{finn2015functional}. In this matching, we used the \mysingleq{Wishart} distance between two correlation matrices, focusing on relevant ROIs in a view. In other words, we evaluated the similarities between a correlation matrix $\bb{A}$ and correlation matrix $\bb{B}$ as
\begin{eqnarray}
\mbox{Similarity} = \mathcal{W}(\bb{B}|T, \bb{A}),
\end{eqnarray}
where we set $T$ to the number of datapoints in BOLD signals.
Note that this measure was not symmetric for $\bb{A}$ and $\bb{B}$. In our context,
we let $\bb{A}$ denote a correlation matrix for a matched subject, whereas $\bb{B}$ for a matching subject (see more details in the caption of Fig.\ref{comparisonrestAccuracy}). In this manner, we evaluated the accuracy of the matching of subjects in both directions for Rest1 and Rest2 (Fig.\ref{comparisonrestAccuracy}a, b). As a result of matching, we found that the accuracy of the matching of subjects was high for those three pairs. In particular, the accuracy reached more than 0.95 for pair 1 and pair 3. For pair 1, the accuracy was 0.96 in Rest1 and 0.97 in Rest2, whereas for pair 3, the accuracy was 0.97 in Rest1 and 0.97 in Rest2. In contrast, for pair 2, the accuracy deteriorated to 0.64 in Rest1 and 0.61 in Rest2.

Regarding the sub-network of the brain, the relevant ROIs for pairs 1-3 were in the following brain regions: prefrontal, motor strip, parietal, and temporal (left) for pair 1; temporal and occipital (right and left) for pair 2; and prefrontal, motor strip, parietal, and temporal (right) for pair 3 (Fig.\ref{comparisonrestBrain}). The result about the relevant brain regions is discussed in more detail in section \ref{interpretation}.

\section{Discussion}
The main contribution of the present paper is to propose a new multiple-view clustering method for correlation matrices without vectorization, 
which provides a useful framework for cluster analysis of FC data. We formulated the proposed method based on Wishart mixture models by means of extension from a single view to multiple views in the nonparametric Bayesian framework. With the whitening procedure for correlation matrices, the usefulness of the present method was demonstrated both in synthetic and real fMRI data. 

In this section, 
first, we discuss the clustering results for the fMRI data from a neuroscientific point of view. Focusing on the relevant brain regions, we compared the results with those of previous studies. In addition, we provide a possible interpretation for connectome fingerprinting using the age of the subjects. Second, we discuss important features of the proposed method, including its limitations.

\subsection{Interpretation of the clustering results} \label{interpretation}
In the task fMRI data, we identified 12 relevant ROIs for the task difference between MOTOR and WM. Regarding the brain regions, these 12 ROIs were located in the motor strip and parietal regions (Fig.\ref{brain}). This result is consistent with previous studies.
In \cite{barch2013function}, it was hypothesized that the MOTOR task was relevant for motor and somatosensory cortices, whereas WM tasks for various brain regions, including the prefrontal, inferior frontal, precentral gyrus, and anterior cingulate. The systematic study \cite{poldrack2017scanning}, which focuses on BOLD signal changes, reports that the supplementary motor cortex is most relevant for the MOTOR task, whereas the middle frontal gyrus for WM. Further, they suggest that the effect size is larger for the MOTOR task, which gives a possible explanation of why only MOTOR-related ROIs are identified in the present study. It is of note that FC's inter-task difference is smaller than the difference in FC between task and resting-state \cite{cole2014intrinsic}, which suggests that the identification of task differences is more challenging than the identification of task and resting differences.

In the resting-state fMRI data, the relevant brain regions for the three pairs 1-3 are identified in the heuristic approach as follows:
Prefrontal, motor strip, parietal, and temporal (left) for pair 1; temporal and occipital (right and left) for pair 2; and prefrontal, motor strip, parietal and temporal (right) for pair 3 (Fig.\ref{comparisonrestBrain}). Further, it was found that the accuracy of the matching of subjects was quite high for pair 1 and pair 3, whereas it deteriorated for pair 2.
These results are consistent with a previous study \cite{finn2015functional}, which demonstrated that the frontoparietal networks could distinguish between individuals, whereas the visual networks less so. Our results are consistent with their analysis of the contributions of individual FCs. The uniqueness of our approach is that we obtained these results in a data-driven manner without assuming connectivity networks in advance.

Moreover, we characterized subject clusters for pair 3, turning our attention to clusters with a large sample size. Among the three pairs in question, the largest overlapping of subjects in clusters was observed in pair 3: between cluster 1 in view 22 of Rest1 and cluster 1 in view 28 of Rest2 (Fig.\ref{comparisonrestcrosstable}). The number of overlaps was ten, whereas 34 subjects belonged to either of these clusters. We consider that the subjects belonging to the same cluster were homogeneous in terms of FC in the relevant ROIs. Ideally, such subjects would belong to the same cluster in both Rest1 and Rest2. However, because of the small sample size of the overlapping subjects in the present study, we did not require this, focusing on the 34 subjects that belonged to either of these clusters. We hypothesized that these subjects had a common attribute other than FC. To test this hypothesis, we characterized these 34 subjects (we refer to it as \mysingleq{group 1}) against the remainder of the subjects (\mysingleq{group 2}), using basic subject information on sex and age. Statistical tests suggest that the sex difference between the two groups was not significant at the 0.05 level (p-value = 0.58 by $\chi^2$ test), whereas the age difference was significant (p= 0.0018 by Mann–Whitney U test). Further analysis of age showed that the subjects in group 1 were older than those in group 2. Considering that the subjects in group 1 were more homogeneous than those in group 2, this result suggests that as age increases, the connectome fingerprint for pair 3 becomes more homogeneous.

\subsection{Features of the proposed method}\label{discussion2}
The proposed method is unique in that it reveals different clustering patterns of objects depending on a specific subset of nodes. An exclusive partition of nodes defines multiple views in which objects are clustered in a specific subset of nodes. Both the partition of nodes and object clustering are simultaneously optimized based on a probability model. Importantly, the structure of the correlation matrix is kept without vectorization. In the simulation study of synthetic data, it was shown that the proposed method outperformed the other methods. It is worth noting that the multiple-view method based on vectorized matrices (i.e., multiple-view clustering based on Gaussian mixture models) failed to recover both view and object cluster structures. A possible reason for this poor performance is as follows. Due to the generating model of synthetic data, the distributions of vectorized features differ significantly. This makes it difficult to fit the data with a single univariate Gaussian distribution, which is assumed in the multiple-view method based on vectorized matrices. This suggests a limitation of this type of multiple-view clustering method, which relies on the vectorization of a correlation matrix.

The proposed method's key assumption is independence between views, which is effectively addressed by means of whitening of correlation matrices. To elucidate the view structure, the idea of the whitening of the correlation matrix is quite new. Combined with a whitening procedure, the proposed method's usefulness is demonstrated in simulation studies both in synthetic and fMRI data.

The tuning parameter in the proposed method is just a few:
the concentration parameter $\alpha$ of CRP for inferring the number of views and the number of clusters. In all applications of the proposed method in the present paper, we set the concentration parameter $\alpha$ to one, which is the default setting for the conventional application of CRP \cite{gelman2013bayesian}.

As regards the computing complexity, the proposed method has a specific property. 
As can be seen in Eq.(\ref{matrixstr}), 
the dimension of correlation matrix decreases when the number of views increases, leading to the reduction of the computing complexity. To take into account this effect, we discuss the computing complexity focusing on the number of views $V$. For simplicity, we fix both the number of subject clusters and the number of iterations without explicitly including them in the evaluation of the computing complexity.
 Further, we assume the same number of nodes over different views, denoting it as $p'$, i.e., $p' = p/V$ where $p$ is the number of all nodes. In this setting, the computing complexity for updating the subject cluster membership is given by $O(nVp'^3)$ where $n$ is the sample size. Note that we set the computing complexity of the determinant of $p' \times p'$ matrix to $p'^3$.
On the other hand, the computing complexity for updating the view membership is given by $O(pV(np' + p'^3))$, where the first term  is due to the evaluation of the mean correlation matrix in a subject cluster. By substituting $p' = p/V$, the overall computing complexity becomes $O((n+p)p^3/V^2)$.
This result implies that the proposed method can efficiently work for a larger number of views, given the fixed number of nodes. Next, we evaluate the computing complexity of the multiple-view clustering based on Gaussian mixture models, which is applied for vectorized matrices. 
For simplicity, we consider the Gibbs sampling version of \cite{tokuda2017multiple}. The computing complexity of such a method is given by $O(nVp'')$, where $p''$ is the number of FC features for each view. In the present context, it becomes that $p'' = (p(p-1)/2)/V$, which leads to the computing complexity 
$O(nVp'') =O(nV(p^2/V)) = O(np^2) $. Hence, the computing complexity increases proportional to the quadratic of the number of nodes, while it is constant for the number of views. Finally, it might be an interesting question to compare the computing complexity between the proposed method and the Gaussian-based method. One may consider a specific case that the number of views $V$ increases proportional to the number of nodes $p$, i.e., $p'=p/V = constant$.  In this case, the computing complexity for the proposed method becomes 
$O(np + p^2)$, which is comparable to that of the Gaussian-based method in terms of the number of nodes $p$. 

It is expected that the present method may provide a useful framework to understand better the brain-related heterogeneity of subjects, such as subtyping of psychiatric disorders, which is currently an active area of research. The task fMRI analysis results in the present study imply that the present method can identify subject groups that differ in FC patterns
in a particular sub-network of the brain. It is worth noting that  in the additional experiment of task-fMRI, the task-difference of MOTOR-WM was not recovered when we imposed the constraint of the single view structure. This suggests the usefulness of the proposed method that explicitly models the underlying multiple view structure. 
However, the reproducibility of subject clusters could not be achieved in the analysis of resting-state fMRI. Nonetheless, this does not rule out the possibility of identifying subtypes of psychiatric disorders using resting-state fMRI, which may stably differ in connectivity patterns. Alternatively, it might be a good option to focus on views with better reproducibility, utilizing scanning resting-state fMRI at least twice.

Finally, we discuss the limitations of the proposed method. First, the proposed method applies only to the correlation matrix, or in general, Gram matrix, which is positive definite. For other types of connectivity matrices, one may consider a probability distribution other than the Wishart distribution. Nonetheless, the incorporation of a multiple-view structure into a probability model is similar. Second, the proposed method provides a cluster solution of objects based on different correlation matrix patterns, but the difference between clusters is not necessarily obvious by visual inspection. This is because a cluster solution is based not on a single element but several correlation matrix elements. To solve this problem, dimension-reduction methods may be useful.
Third, the proposed method is based on the assumption that the correlation matrix may follow a Wishart distribution in each cluster block. This assumption may not hold for real FC data. In that case, a yielded cluster should be considered just as a mixture component. Note that such a situation may in general arise in clustering using mixture models. Since it is expected that a non-Wishart cluster may be fitted by several Wishart mixture components, the merging of several components in close proximity can theoretically provide information on the true cluster structure. However, we did not address this issue in the present paper because of the complication of the problem.

\section{Conclusion}
For the analysis of fMRI data, 
we proposed a novel multiple-view clustering method for correlation matrices. The main feature of this method is to optimally partition nodes into several exclusive groups (views), whereas objects are clustered using only relevant nodes for each view. The assumption of independence between views is addressed by means of  whitening correlation matrices. 
It is of particular note that the structure of correlation matrices is preserved, as evaluated using the Wishart distribution. The partitioning of nodes and clustering of objects, including the number of views and clusters in each view, is simultaneously performed without a tuning parameter. The power of this novel method is shown in a simulation study of synthetic data, which outperforms other clustering methods.

The proposed method provides a useful framework for the analysis of FC data. In the analysis of task fMRI data, it was shown that the proposed method identifies task differences in a data-driven manner. In addition, in the analysis of resting-state fMRI data, the proposed method identifies ROI networks related to connectome fingerprints in a data-driven manner. It is expected that the proposed method may provide a new framework for clustering problems in various contexts, such as the subtyping of psychiatric disorders.


\section*{Acknowledgements}
This research was (partially) conducted as a contract project supported by AMED under Grant Number JP20dm0307002 (TT), JP18dm0307008 (JY, TT),
JP19dm0307009 (OY). Data were provided [in part] by the Human Connectome Project, WU-Minn Consortium (Principal Investigators: David Van Essen and Kamil Ugurbil; 1U54MH091657) funded by the 16 NIH Institutes and Centers that support the NIH Blueprint for Neuroscience Research; and by the McDonnell Center for Systems Neuroscience at Washington University. We thank Dr. Hiroshi Morioka at RIKEN for his useful comment on a whitening procedure. 

\bibliographystyle{unsrt}

\begin{figure}[!]
\begin{flushleft}
\includegraphics[scale=0.18, trim= 200 20 0 0]{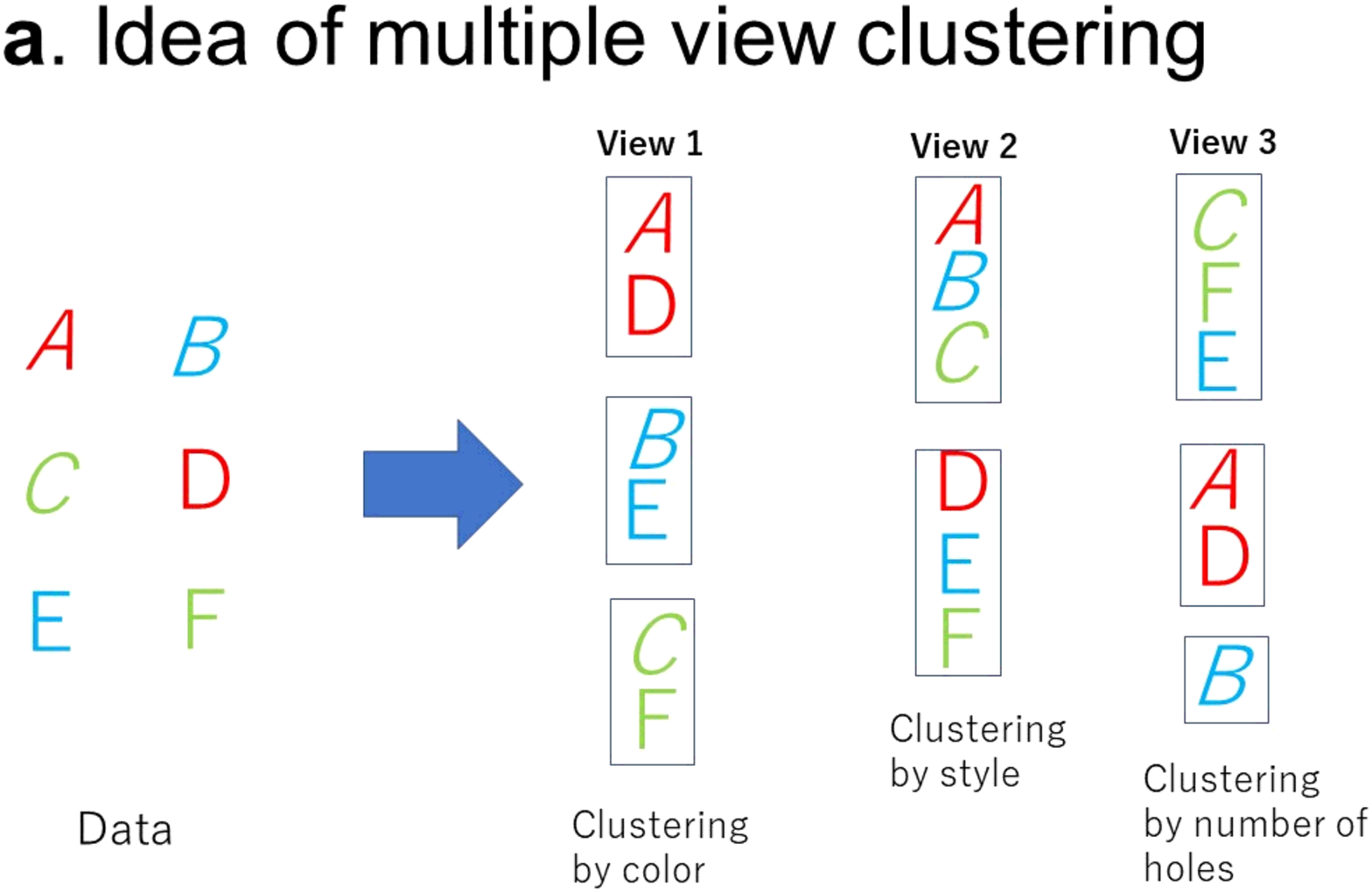}
\includegraphics[scale=0.18, trim= 50 40 200 0]{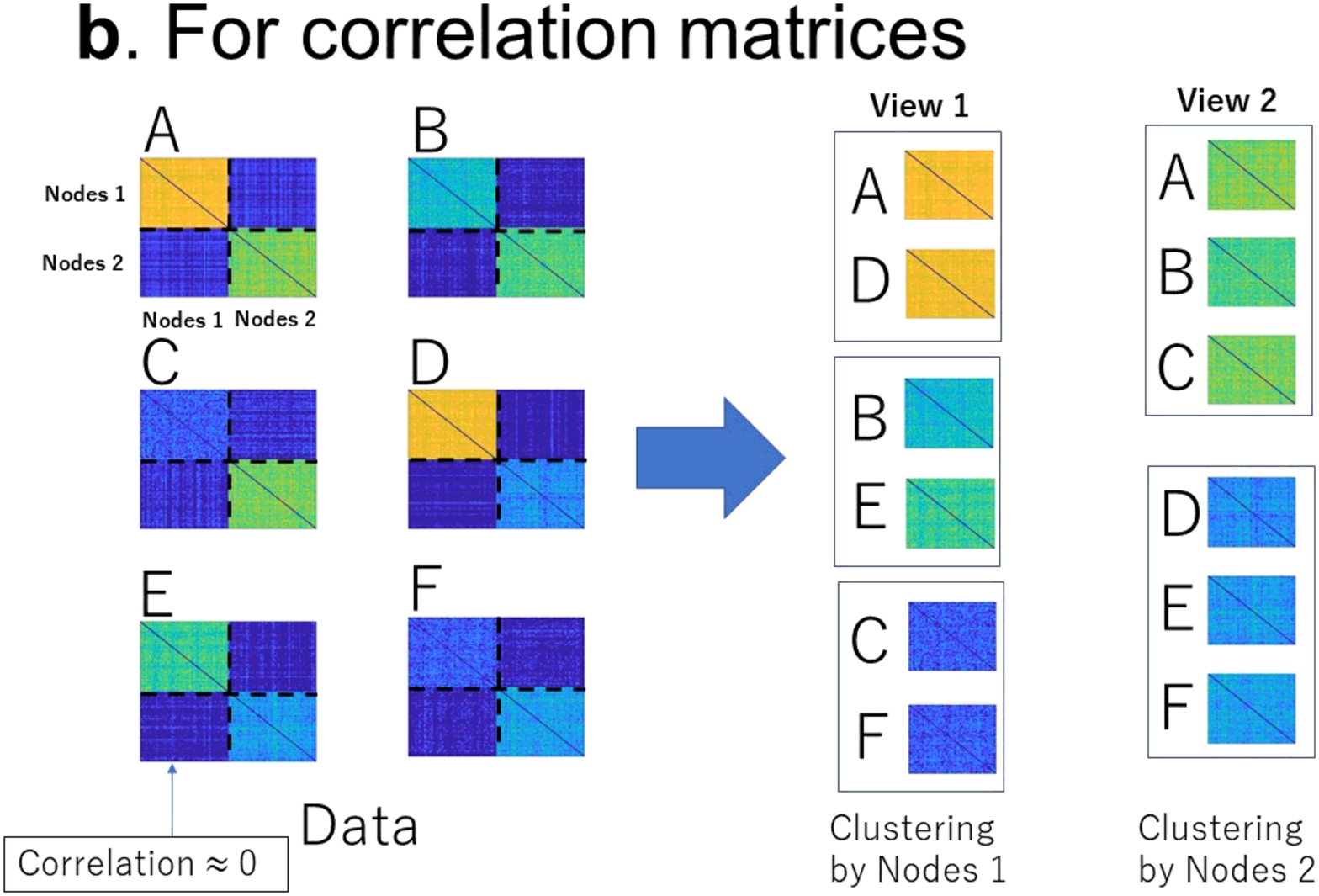}

\vspace{5mm}

\includegraphics[scale=0.18, trim= 50 0 0 0]{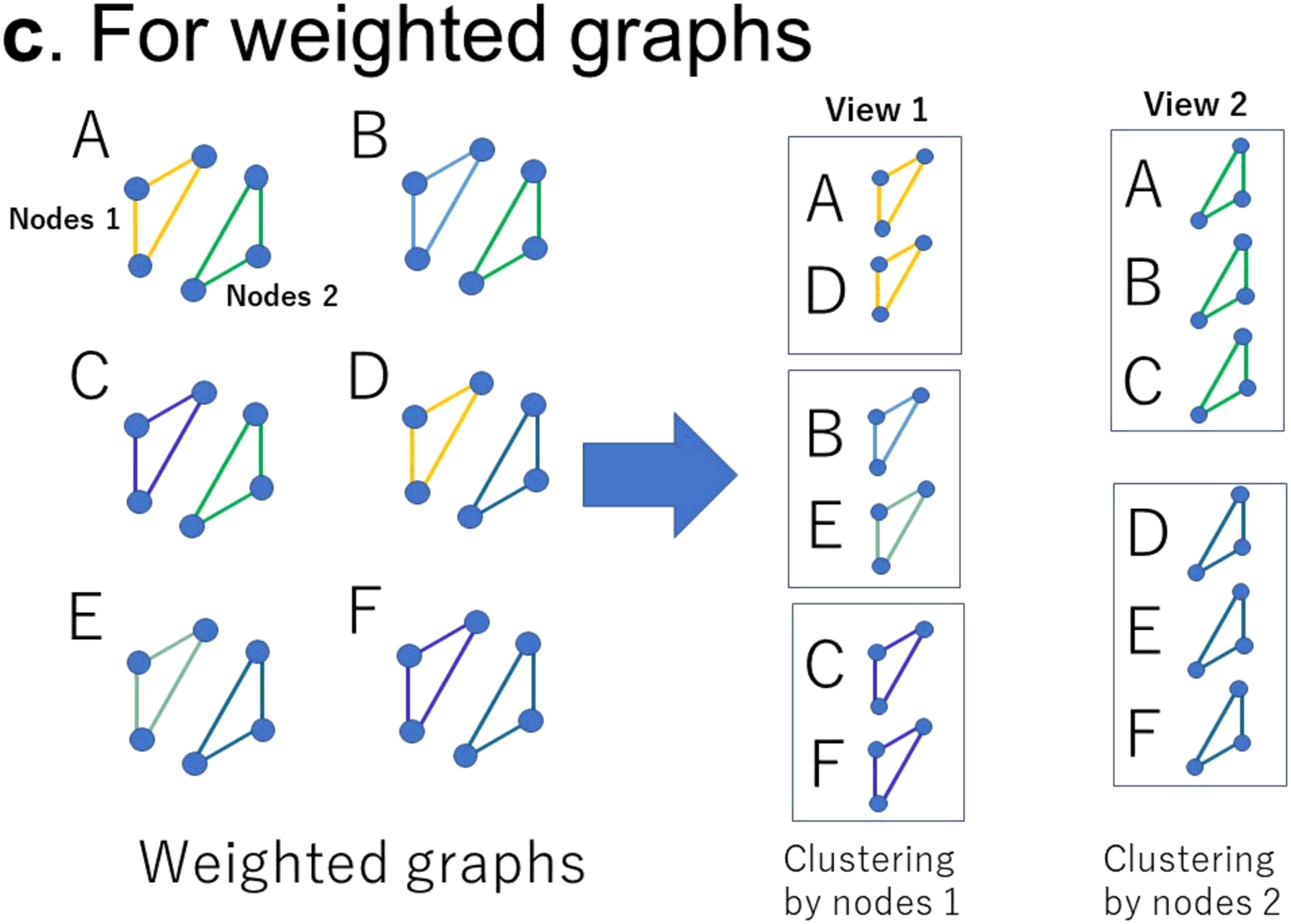}
\end{flushleft}
\caption{\small Illustration of multiple-view clustering. Panel a: Basic idea of 
multiple-view clustering. In this toy example, we consider clustering of six letters, focusing on a particular feature. Views 1-3 show clustering results (letters in the same square belong to the same cluster), focusing on color, style and the number of holes, respectively. 
Panel b: Multiple-view clustering for correlation matrices. We consider clustering of six correlation matrices A-F with the matrix size $100 \times 100$. 
Color denotes Pearson's correlation (the diagonal part is set to zero).
Note that in the conventional terminology, the correlation matrix denotes the association between two variables. However, here, instead of \mysingleq{vairable}, we use the terminology \mysingleq{node}, which is more appropriate as a synonym for ROI in the present paper. 
If we focus on the first 50 nodes (nodes 1), we have a two-cluster solution as denoted in view 1. 
On the other hand, if we focus on the last 50 nodes (nodes 2), we have a three-cluster solution as in view 2. Note that we assume independent networks nodes 1 and nodes 2 (the between-network correlation is zero). Also, we assume an exclusive partition of nodes, where each node belongs to only a single view.
Panel c: Illustration of Panel b in terms of undirected weighted graphs. For simplicity, we consider three nodes for nodes 1 and nodes 2, respectively. The strength of connectivity (weight) between two nodes is denoted by color in an edge. Absence of edge denotes zero weight. 
}
\label{Illustration}
\end{figure}

\begin{figure}[!]
\begin{center}
\includegraphics[scale=0.45]{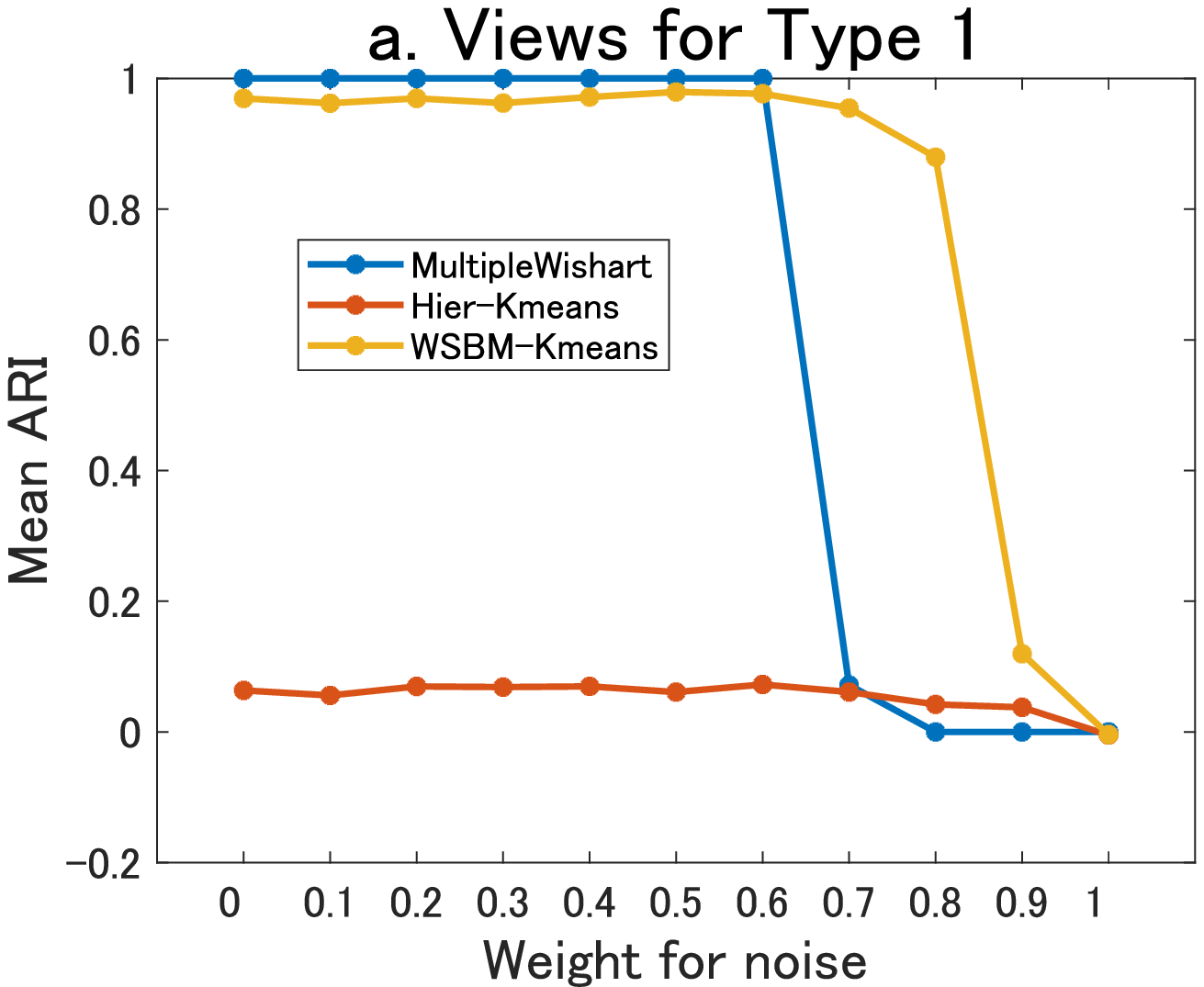}
\includegraphics[scale=0.45]{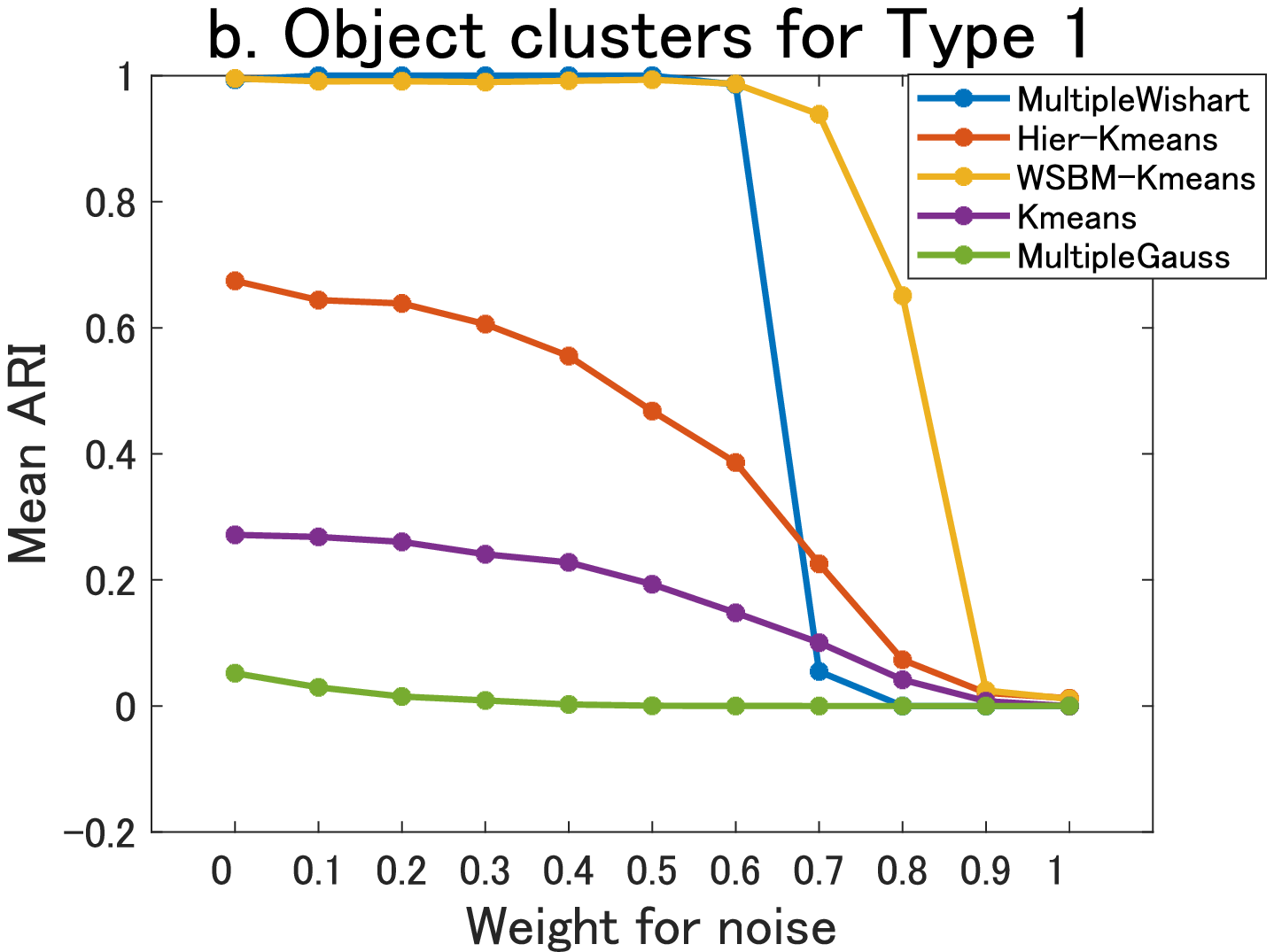}
\includegraphics[scale=0.45]{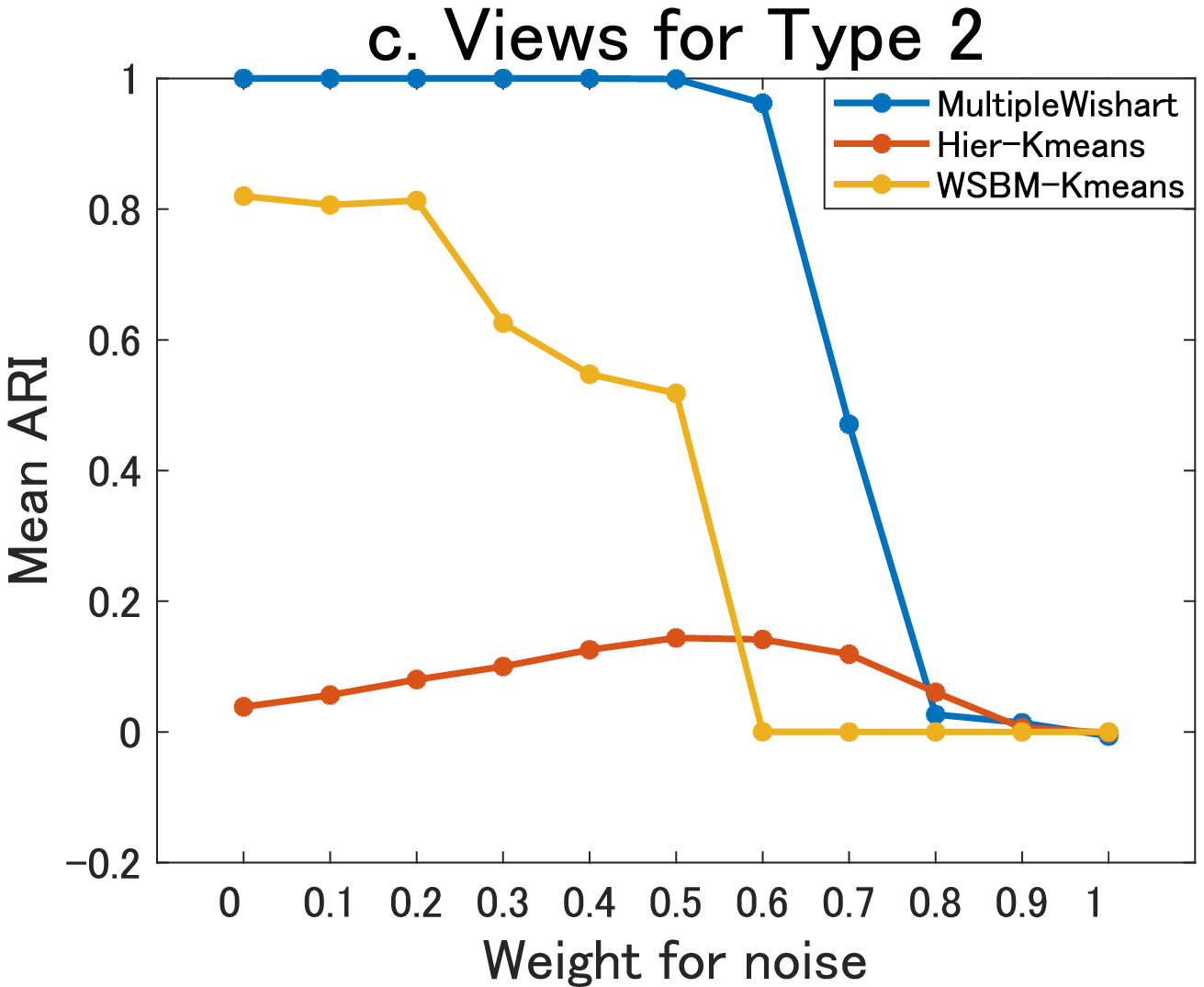}
\includegraphics[scale=0.45]{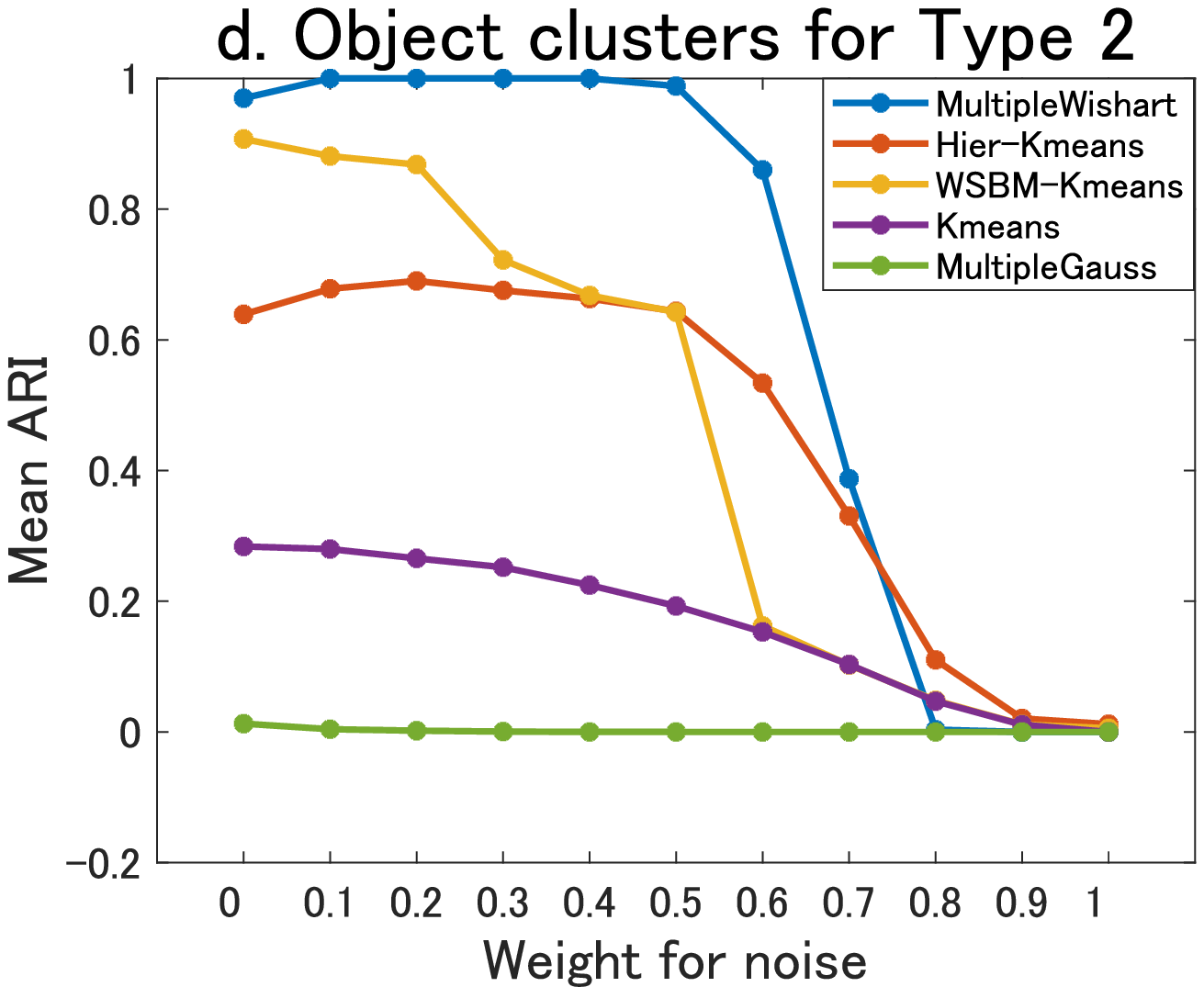}
\end{center}
\caption{Results of simulation study on synthetic data. Panels a, b: Recovery of views
and recovery of object clusters for Type 1 data (background correlation 0 between views) for the proposed method (\mysingleq{MultipleWishart}), hierarchical + K-means (\mysingleq{Hier-Kmeans}), 
community detection + K-means (\mysingleq{WSBM-Kmeans}), K-means (\mysingleq{K-means}), and multiple clustering based on Gaussian mixture models (\mysingleq{MultipleGauss}). Panels c, d: Recovery of views
and recovery of object clusters for Type 2 data (background correlation 0.2 between views). For all panels, the horizontal axis denotes a noise weight $w$ in Eq.(\ref{noiseratio}). The recovery is evaluated by mean Adjusted Rand Index over 100 replications of datasets for each configuration. Note that recovery of views is not applicable for the method of K-means, which does not yield a solution for the view structure. Also, it is not applicable for the multiple view clustering based on Gaussian mixture models, because it yields views based on matrix elements, rather than nodes.
}
\label{simres}
\end{figure}

\begin{figure}[!]
\begin{center}
\includegraphics[scale=0.60]{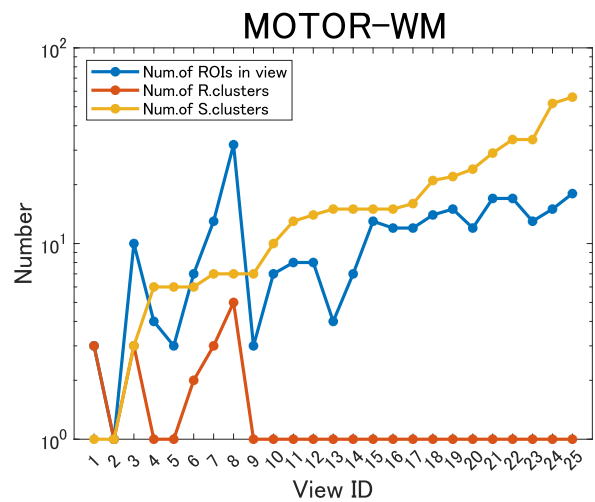}
\end{center}
\caption{Summary of clustering results for dataset MOTOR-WM. 
The horizontal axis denotes view ID, whereas the vertical axis (logarithm scale) the relevant figures for clustering: the number of ROIs in view (blue); the number of ROI clusters in view (red); the number of subject clusters in view (yellow). Note that in each panel, views are sorted in the ascending order
of the number subject clusters in a view.}
\label{summary1}
\end{figure}

\begin{figure}[!]
\begin{center}
\includegraphics[scale=0.60]{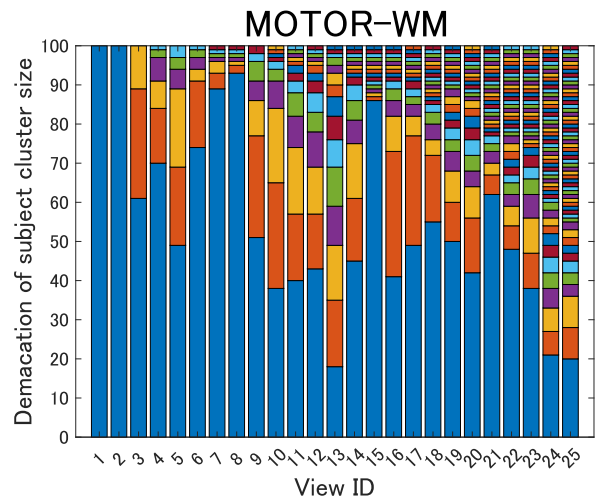}
\end{center}
\caption{Summary of subject cluster size for dataset MOTOR-WM. The horizontal axis denotes view ID, whereas the vertical axis denotes the demarcation of subject clusters. In each color bar, subjects are sorted in the ascending order of subject cluster ID. On the other hand, subject cluster ID is in advance sorted in the descending order of its cluster size. Cluster differences are denoted by color, but note that the same color may be used for different clusters apart.}
\label{summary2}
\end{figure}

\begin{figure}[!]
\begin{center}
\includegraphics[scale=0.60]{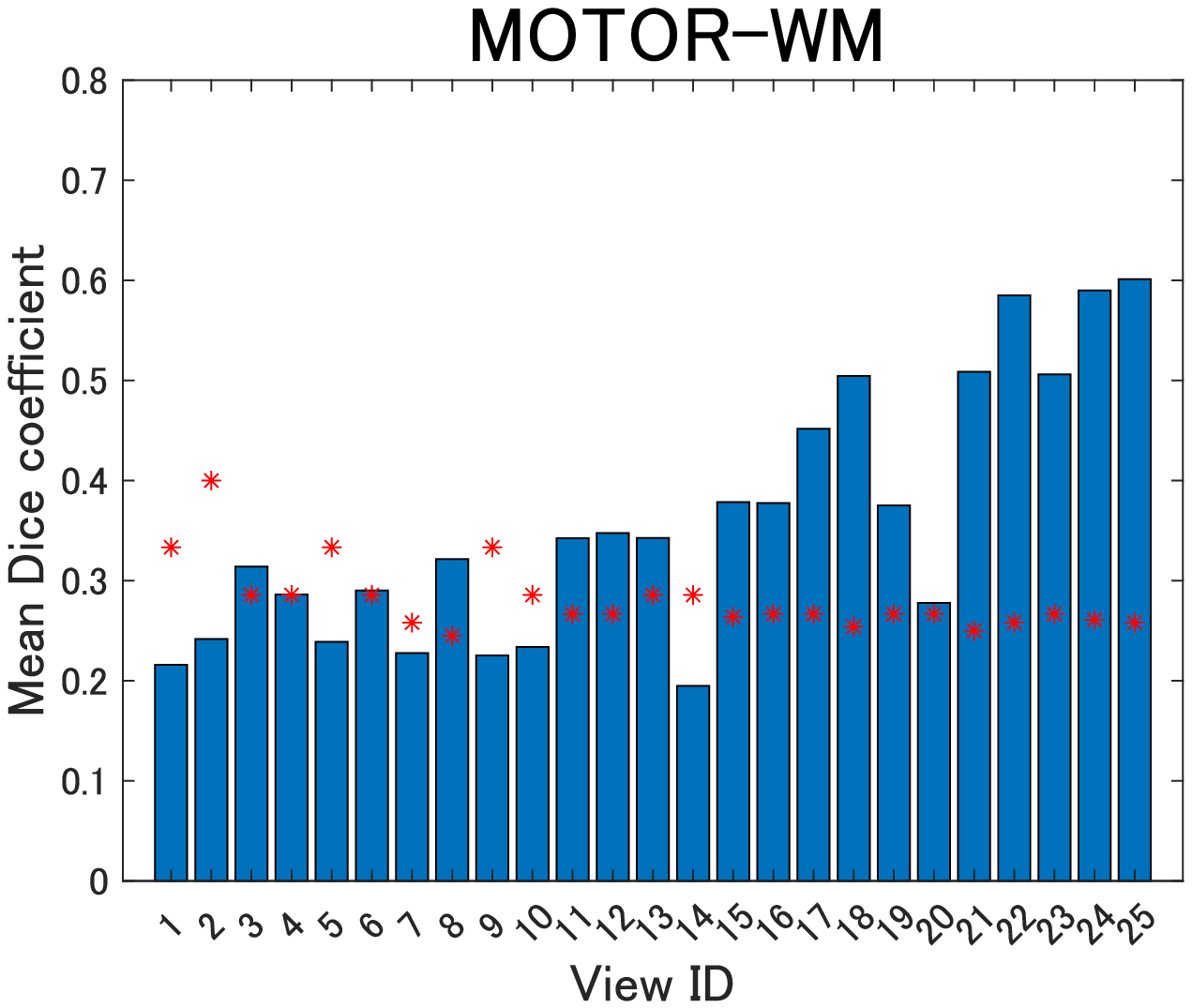}
\end{center}
\caption{Stability of view memberships for dataset MOTOR-WM. Mean Dice coefficient of view memberships between the optimal model and the remainder of the 30 best models (blue bar). View matching among these models is performed by means of maximization of Dice coefficient. Red dots denote the 0.95 quantile when permutation test is performed (1000 permutations). }
\label{stability}
\end{figure}

\begin{figure}[!]
\begin{center}
\includegraphics[scale=0.60]{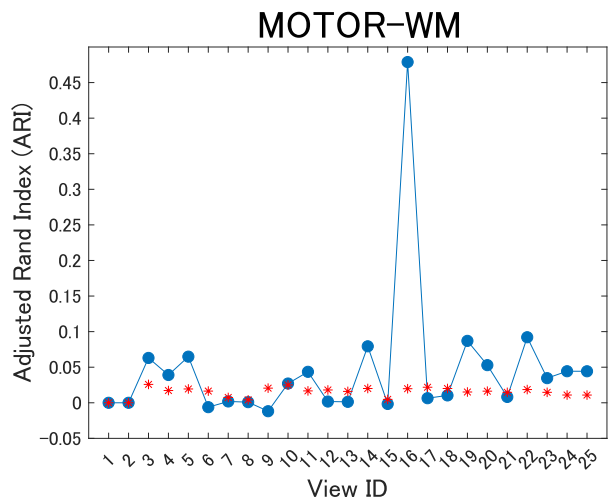}
\end{center}
\caption{Recovery of the task difference
for dataset MOTOR-WM. The horizontal axis denotes view ID, whereas the vertical axis adjusted Rand index between labels of tasks and labels of subject cluster membership. Red points denote
the 0.95 quantile when permutation test is performed for cluster memberships (1000 permutations).}
\label{recovery}
\end{figure}

\begin{figure}[!]
\begin{center}
\includegraphics[scale=0.6]{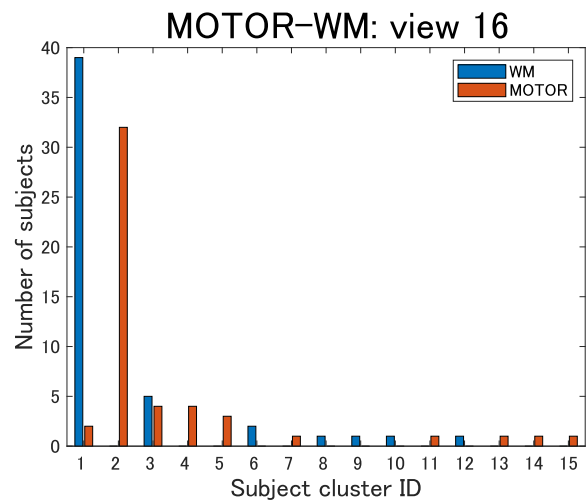}
\end{center}
\caption{Task-wise distributions of subjects for view 16 in dataset MOTOR-WM. The horizontal axis denotes subject cluster ID in the view. The vertical axis denotes the number of subjects in a cluster, which is summarized in a task-wise manner in different colors.
}
\label{subjectcluster}
\end{figure}

\clearpage
\begin{figure}[!]
\begin{center}
\includegraphics[scale=0.6]{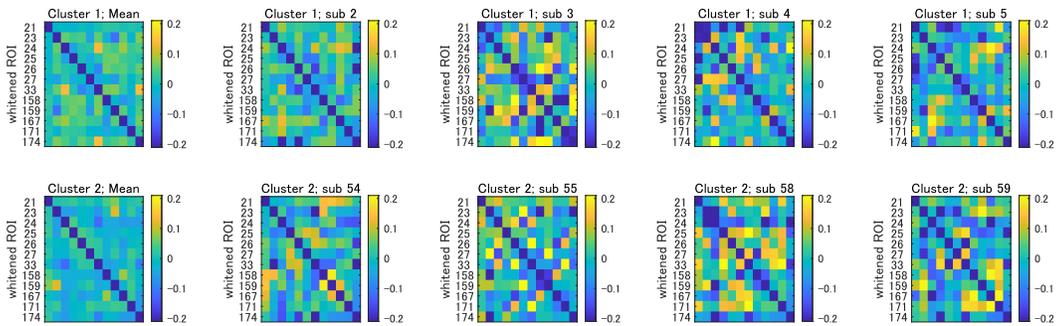}
\end{center}
\caption{Heatmaps of correlation matrices for view 16 in dataset MOTOR-WM. The first column denotes mean correlation matrices over subjects for subject cluster 1 (top) and subject cluster 2 (bottom). For the second to the fifth columns, a correlation matrix is displayed for several subjects in the corresponding subject cluster. Both horizontal and vertical axes denote ROI ID in view 16. Note that these ROIs denote 
\mysingleq{whitened} ROIs after the pre-processing of whitening.}
\label{white}
\end{figure}

\clearpage
\begin{figure}[!]
\begin{center}
\includegraphics[scale=0.5]{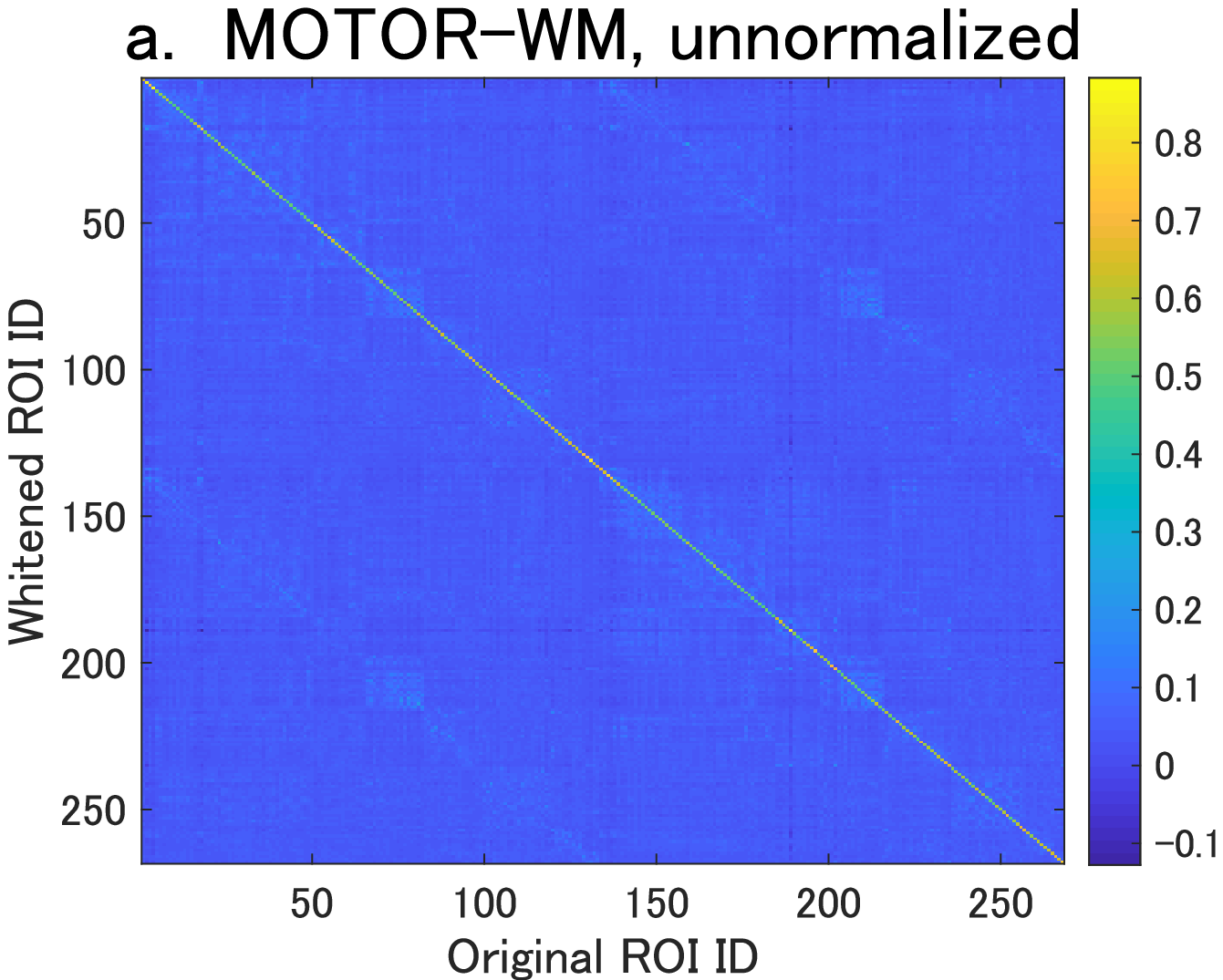}
\includegraphics[scale=0.5]{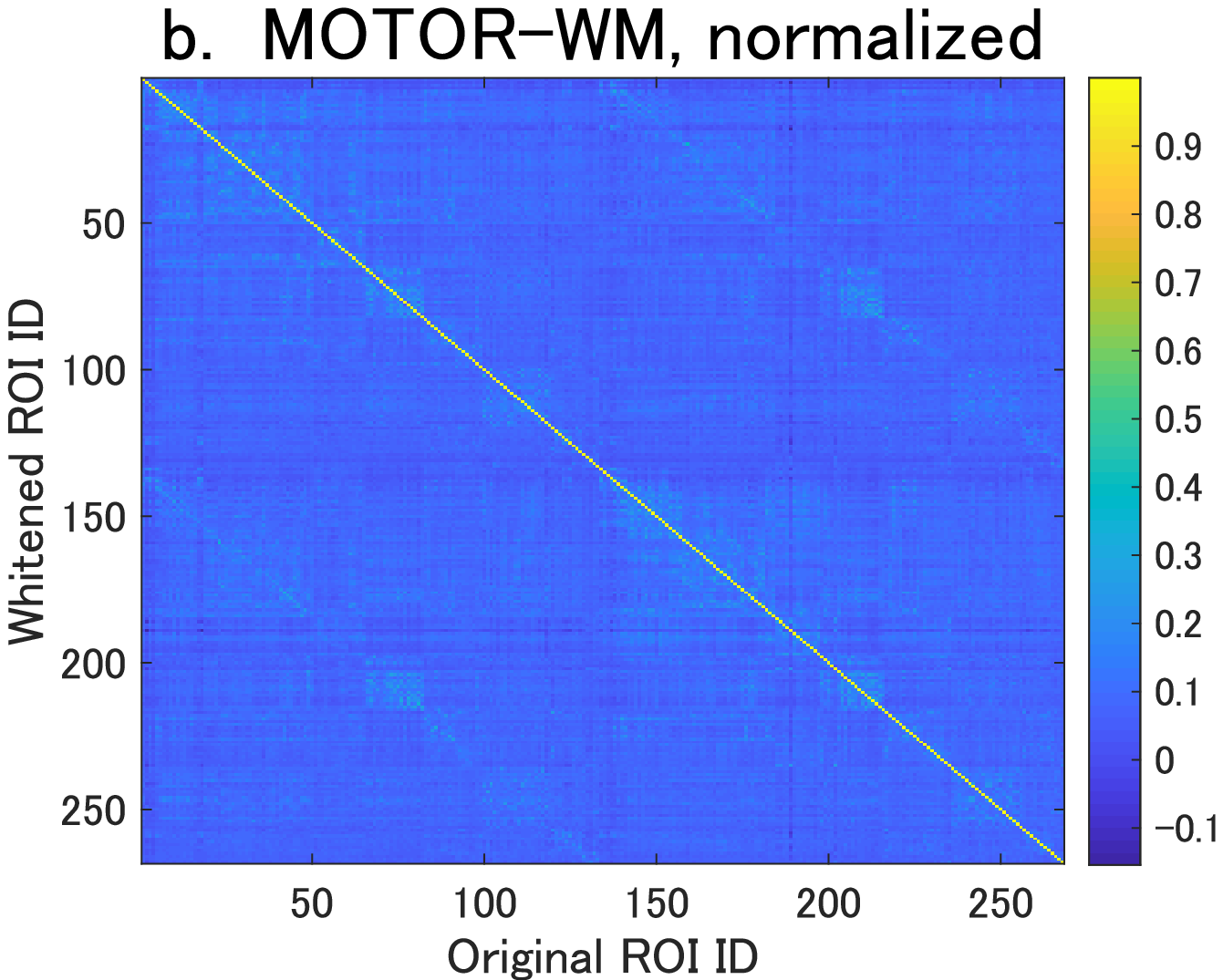}
\end{center}
\caption{Cross-correlations between original ROIs and whitened ROIs for dataset MOTOR-WM. Panel a: The horizontal axis denotes the original ROIs, while the vertical axis the whitened ROIs. Panel b: normalized cross-correlations. The normalization is performed as follows: Values in each row are scaled (i.e., divided) by the maximum values in the row. It is observed that the diagonal part takes the maximum value (one) in each row.}
\label{correspandencewhite}
\end{figure}

\begin{figure}[!]
\begin{center}
\includegraphics[scale=0.6]{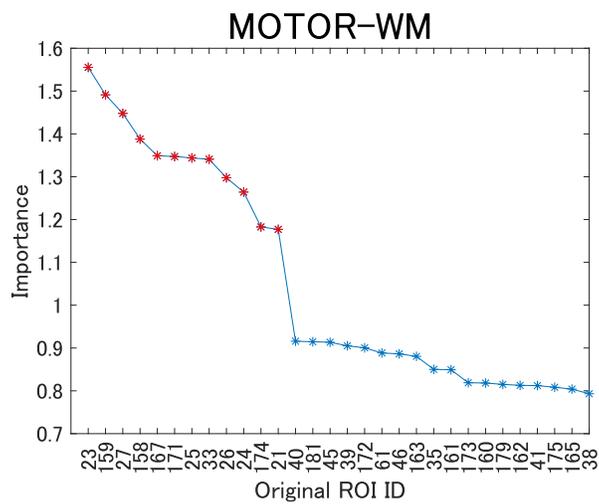}
\end{center}
\caption{Importance of original ROIs for view 16 in dataset MOTOR-WM. The horizontal axis denotes the original ROI ID, while the vertical axis denotes the value of importance that is evaluated by Eq.(\ref{imp}). The top 20 ROIs are displayed, which are sorted in the descending order of the value of importance. Red points denote those ROIs identified in the heuristic approach as well. 
}
\label{rig}
\end{figure}

\begin{figure}[!]
\begin{flushleft}
\includegraphics[scale=0.35, trim=50mm 50mm 0mm 0mm]{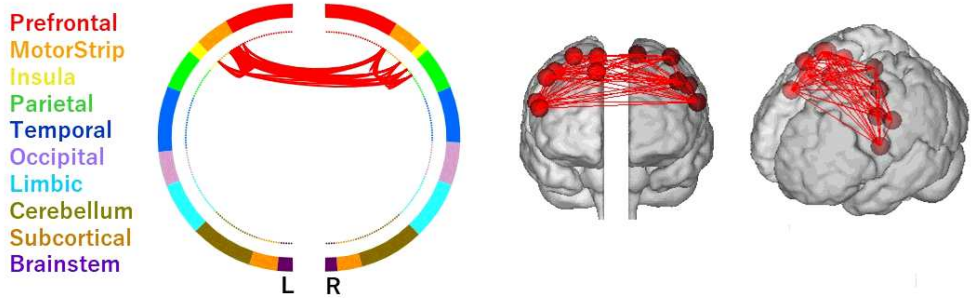}
\end{flushleft}
\caption{Relevant brain regions for the task difference in view 16 of dataset MOTOR-WM. Connecting lines are drawn among 12 relevant ROIs 21, 23, 24, 25, 26, 27, 33, 158, 159, 167, 171 and 174 in Shen parcellation. The visualization is performed by the image analysis software \textit{BioImage Suite} \cite{papademetris2006bioimage}. Note that we used the default setting of view image in which the subject faces the screen (hence right is on the left).
}
\label{brain}
\end{figure}

\begin{figure}[!]
\begin{center}
\includegraphics[scale=0.50]{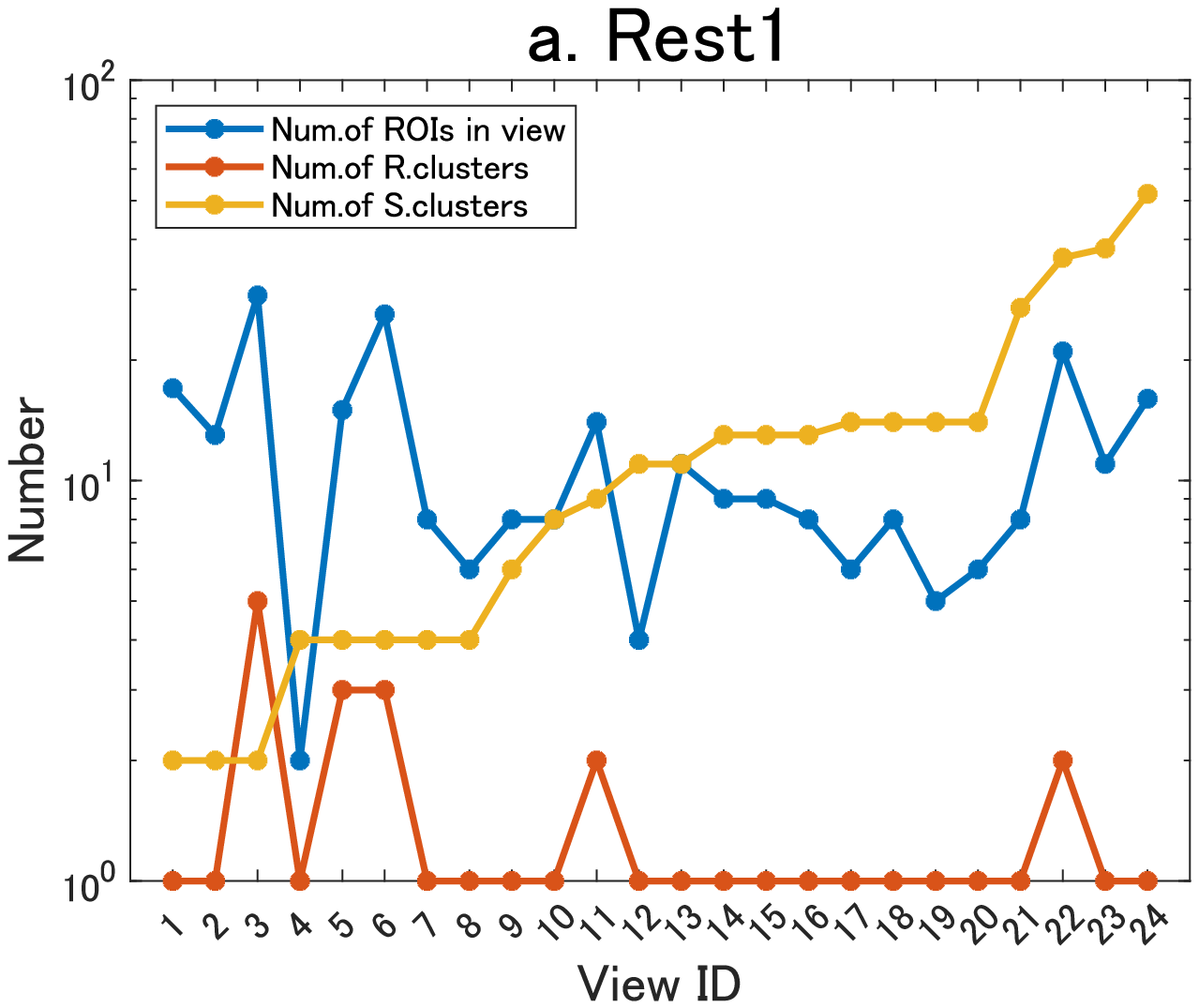}
\includegraphics[scale=0.50]{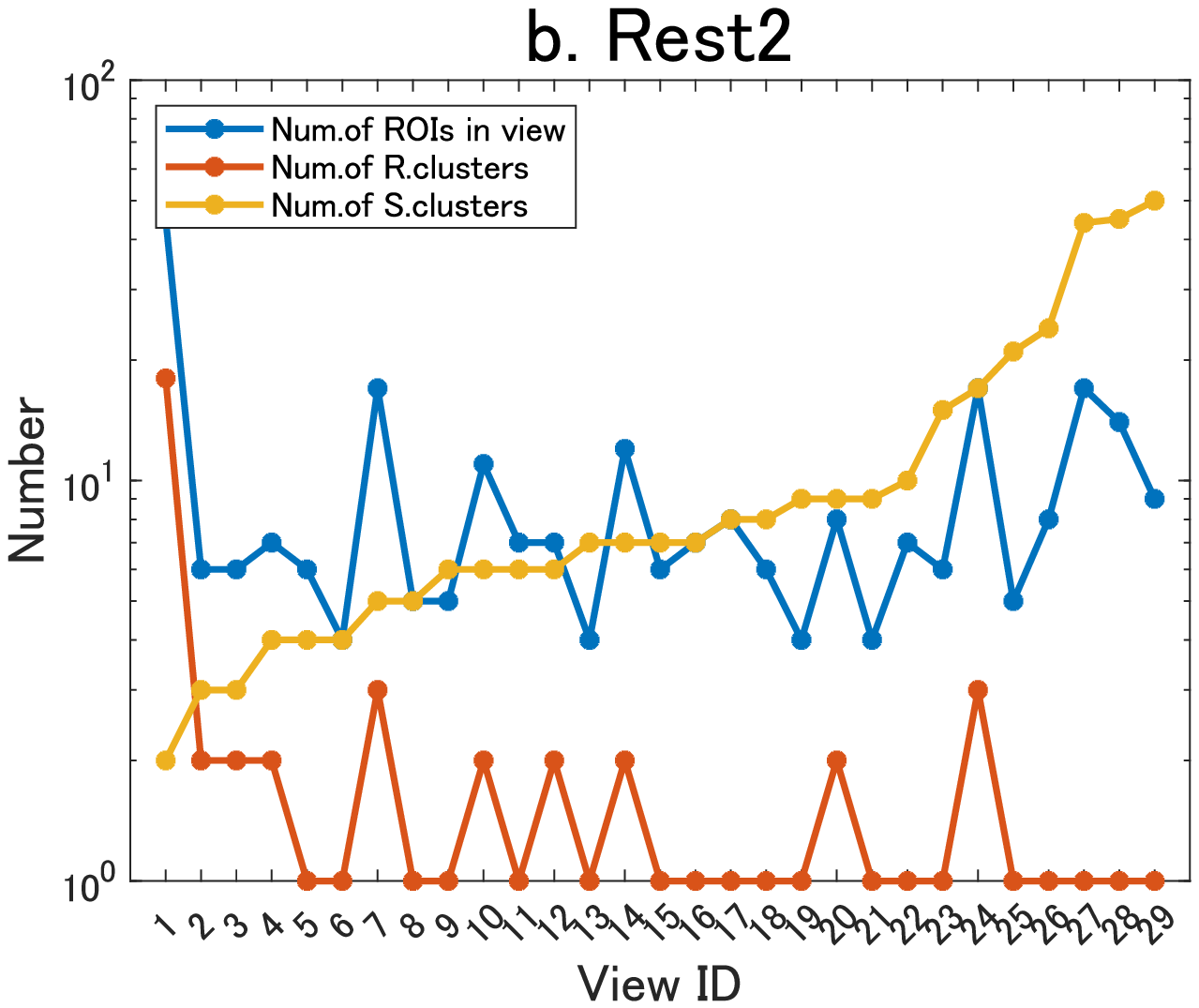}
\end{center}
\caption{Summary of clustering results for resting fMRI data: Rest1 (Panel a); Rest2 (Panel b). The horizontal axis denotes view ID, whereas the vertical axis (logarithm scale) the relevant figures for clustering: the number of ROIs in view (blue); the number of ROI clusters in view (red); the number of subject clusters in view (yellow). Note that in each panel, views are sorted in the ascending order
of the number of subject clusters in a view.}
\label{summary1rest}
\end{figure}

\begin{figure}[!]
\begin{center}
\includegraphics[scale=0.50]{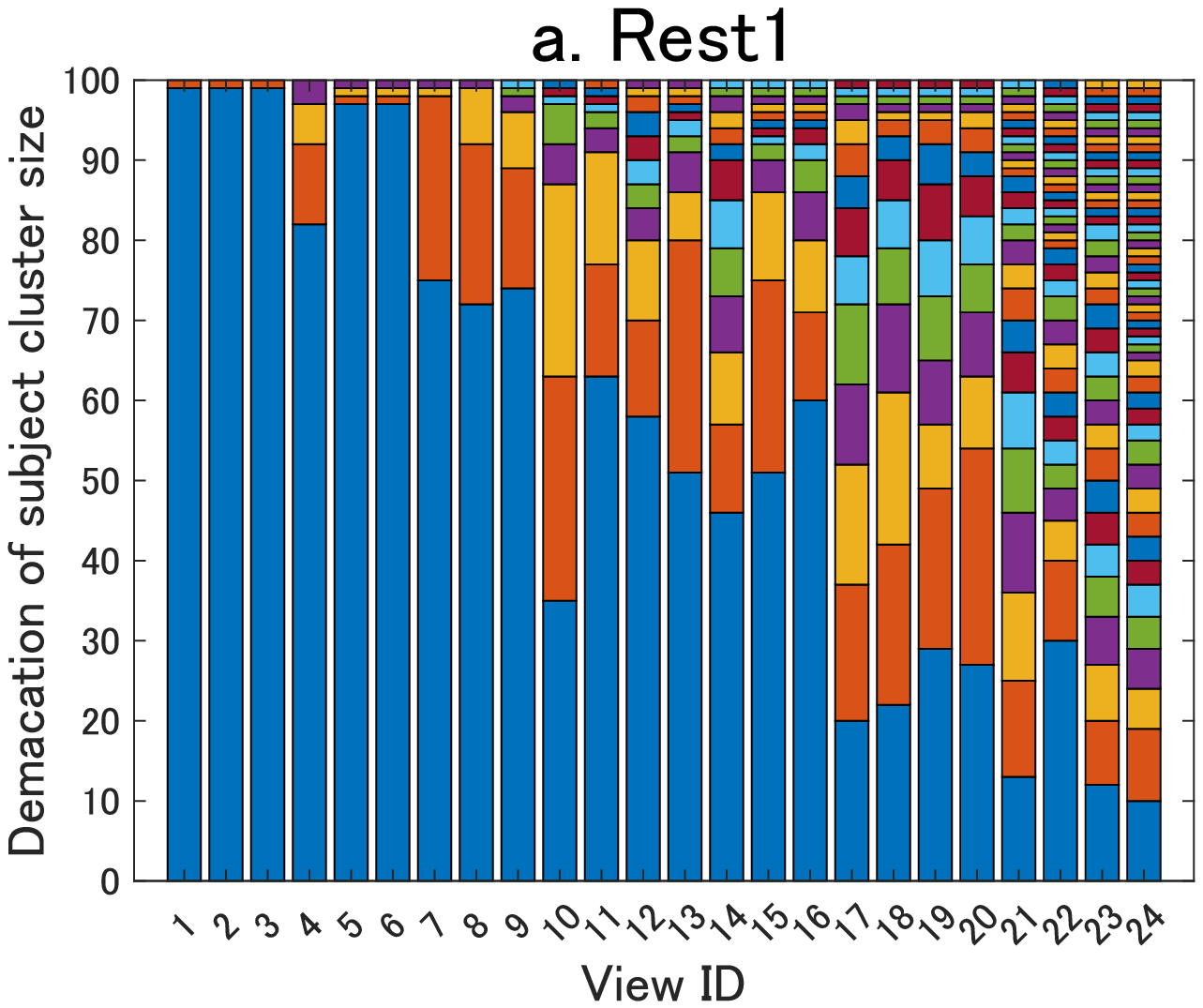}
\includegraphics[scale=0.50]{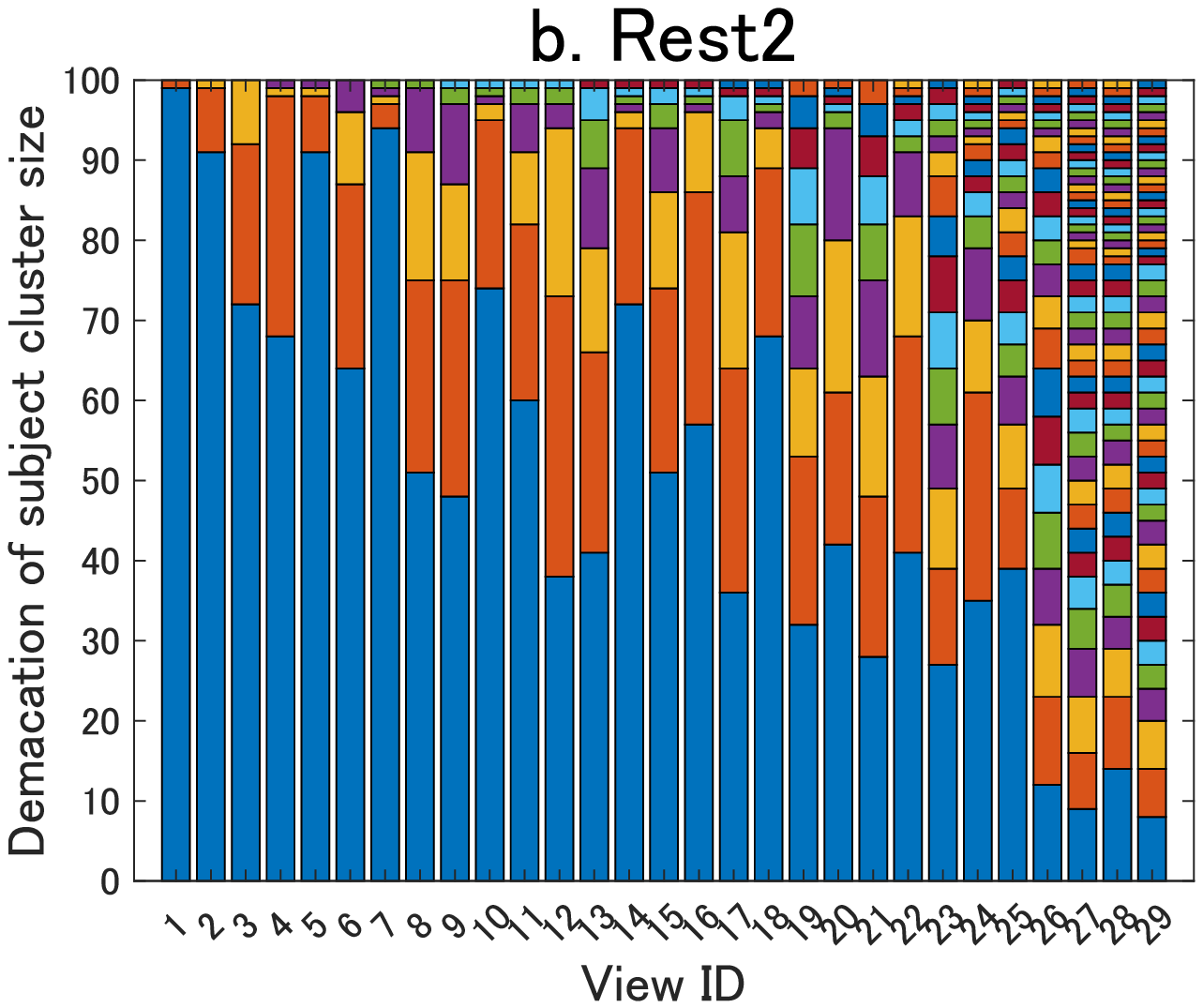}
\end{center}
\caption{Summary of subject cluster size for Rest1 (Panel a); Rest2 (Panel b). 
The horizontal axis denotes view ID, whereas the vertical axis denotes the demarcation of subject clusters. In each color bar, subjects are sorted in the ascending order of subject cluster ID. On the other hand, subject cluster ID is in advance sorted in the descending order of its cluster size. Cluster differences are denoted by color, but note that the same color may be used for different clusters apart.}
\label{summary2rest}
\end{figure}

\begin{figure}[!]
\begin{center}
\includegraphics[scale=0.50]{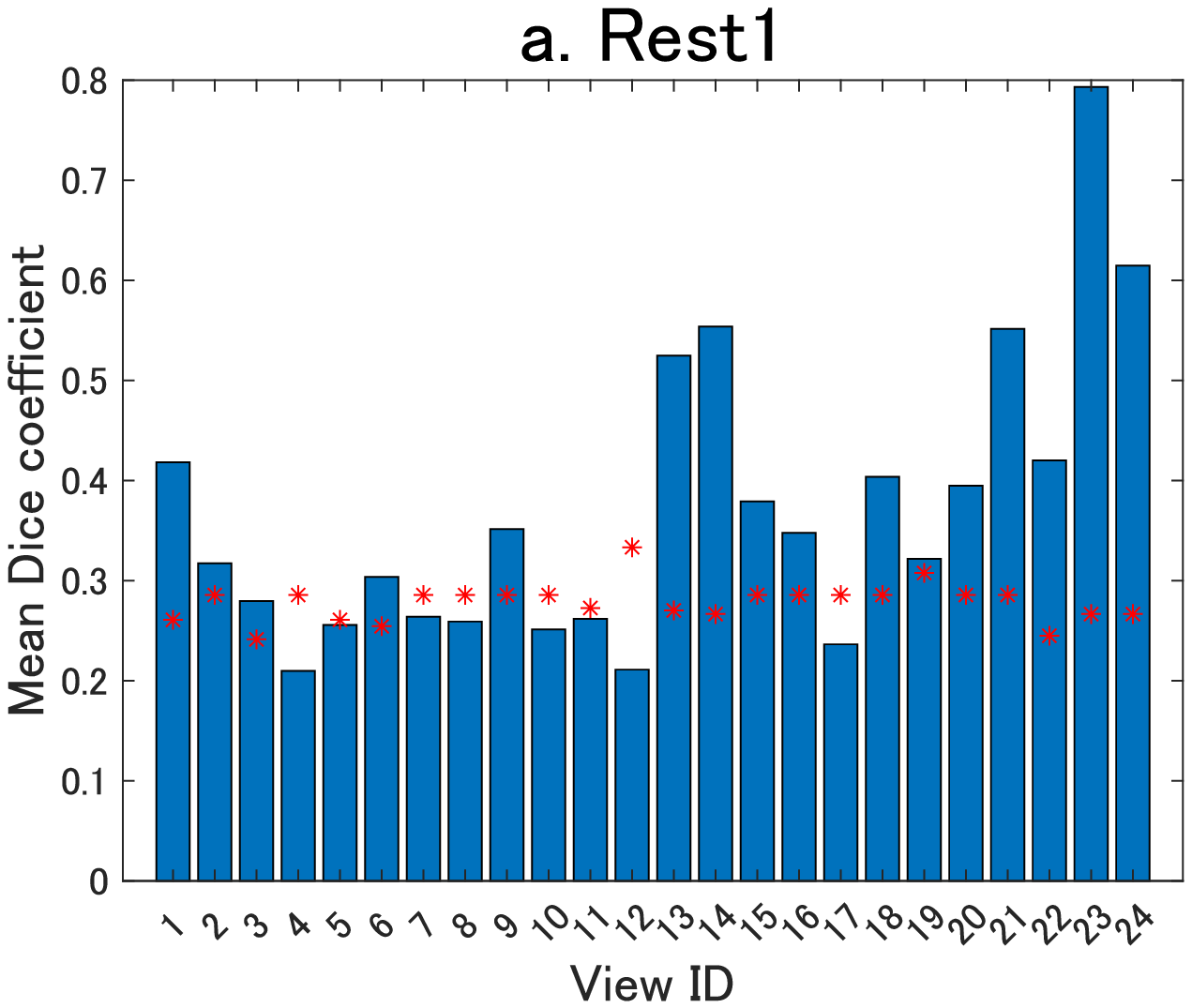}
\includegraphics[scale=0.50]{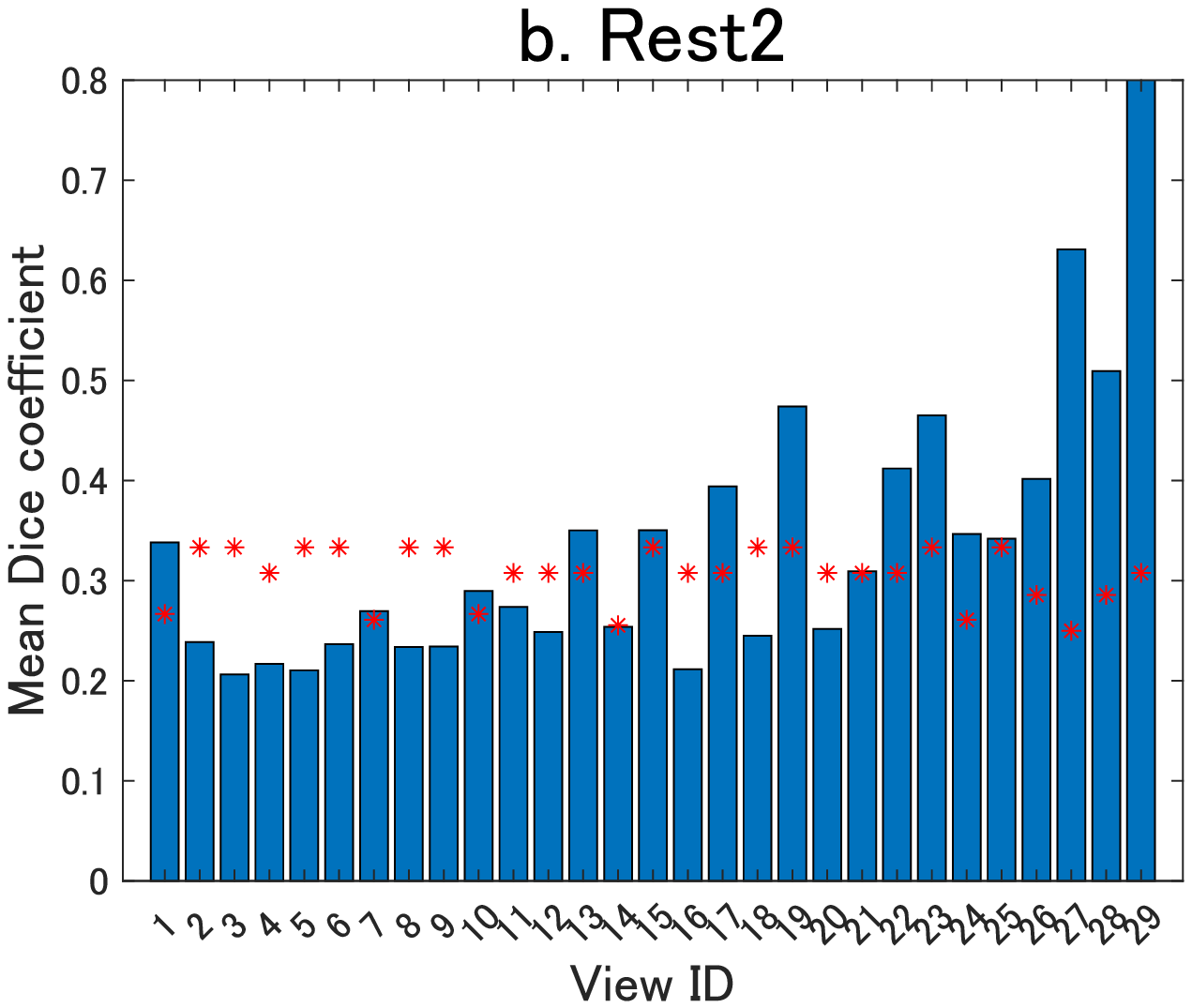}
\end{center}
\caption{Stability of view memberships for Rest1 (Panel a); Rest2 (Panel b). 
Mean Dice coefficient of view memberships between the optimal model and the remainder of the 30 best models (blue bar). View matching among these models is performed by means of maximization of Dice coefficient. Red dots denote the 0.95 quantile when permutation test is performed (1000 permutations). }
\label{stabilityrest}
\end{figure}

\begin{figure}[!]
\begin{center}
\includegraphics[scale=0.50]{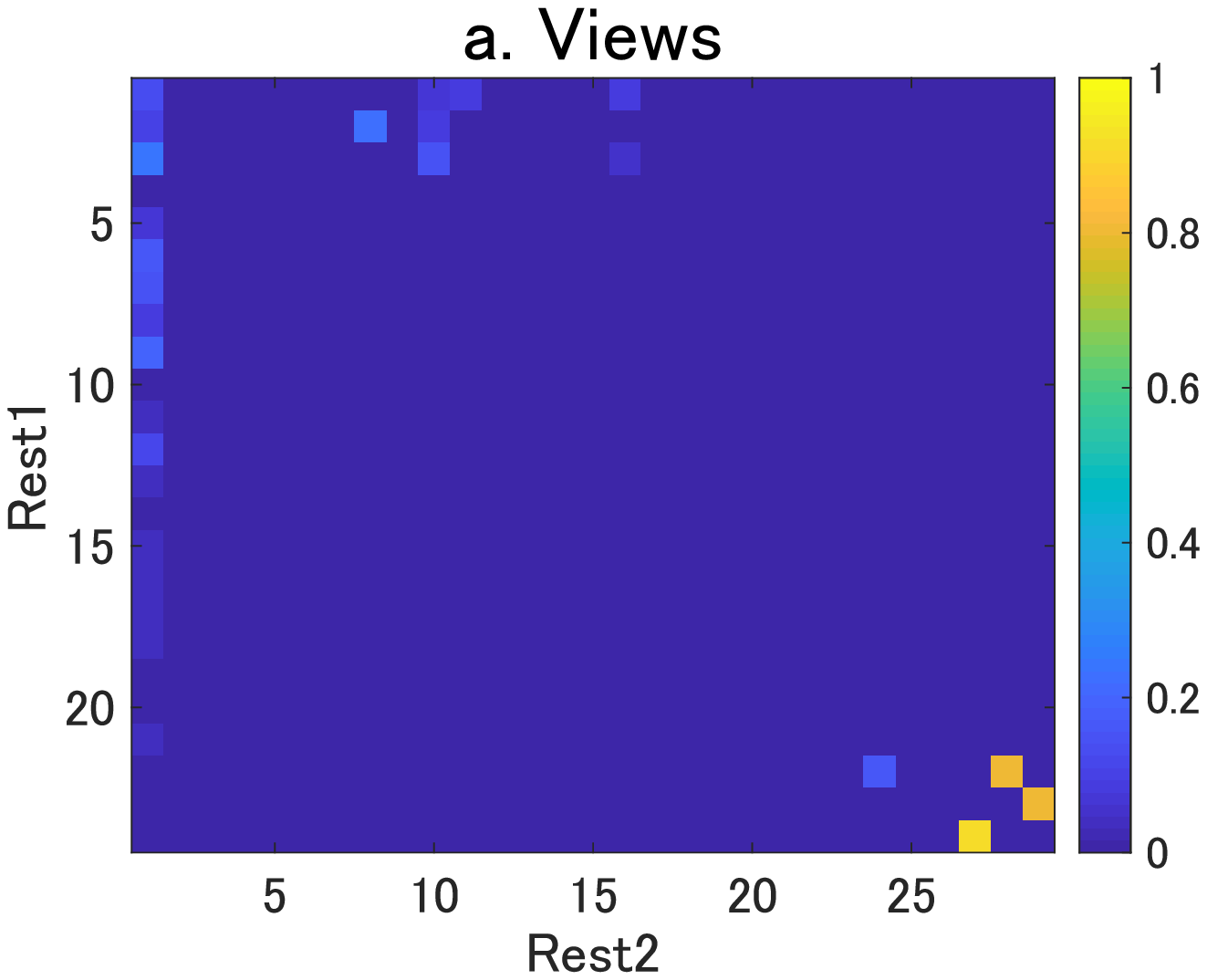}
\includegraphics[scale=0.50]{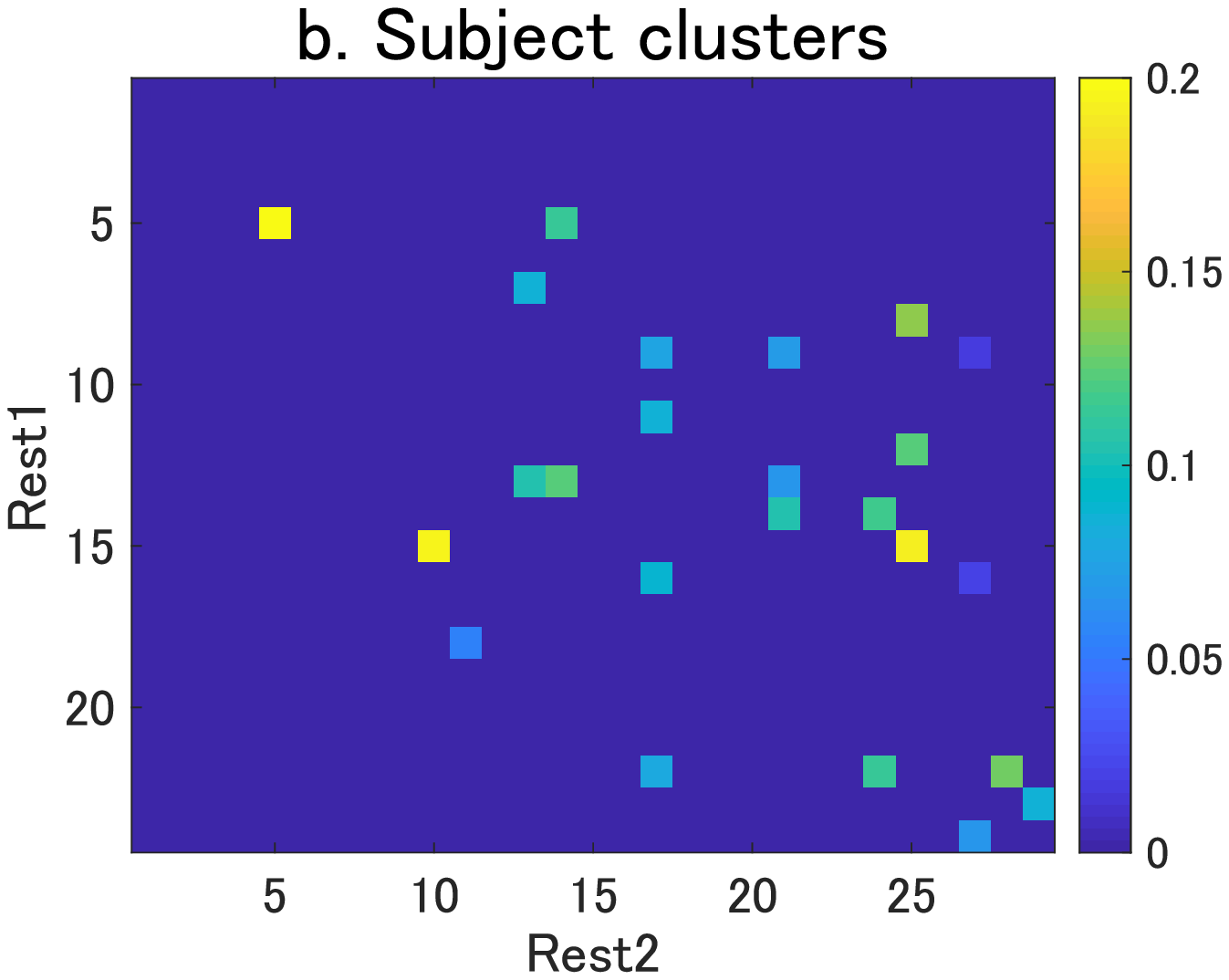}
\end{center}
\caption{Agreement between Rest1 and Rest2: Panel a: Agreement of view memberships between Rest1 and Rest2 evaluated by Dice coefficients.
Only significant Dice coefficients are displayed at significant level 0.05 by permutation test with false discovery rate (FDR) adjustment. Panel b: Agreement of subject cluster memberships between Rest1 and Rest2 evaluated by adjusted Rand index (ARI). Only significant ARIs are displayed at significant level 0.05 by permutation test with FDR adjustment. 
It is of note that the results related to view 1, 2, and 3 in Rest1, and to view 1 in Rest2 are removed, because for those views, values of ARI are inflated due to extreme unbalanced distribution of subjects in subject clusters. }
\label{comparisonrest}
\end{figure}

\begin{figure}[!]
\begin{center}
\includegraphics[scale=0.50]{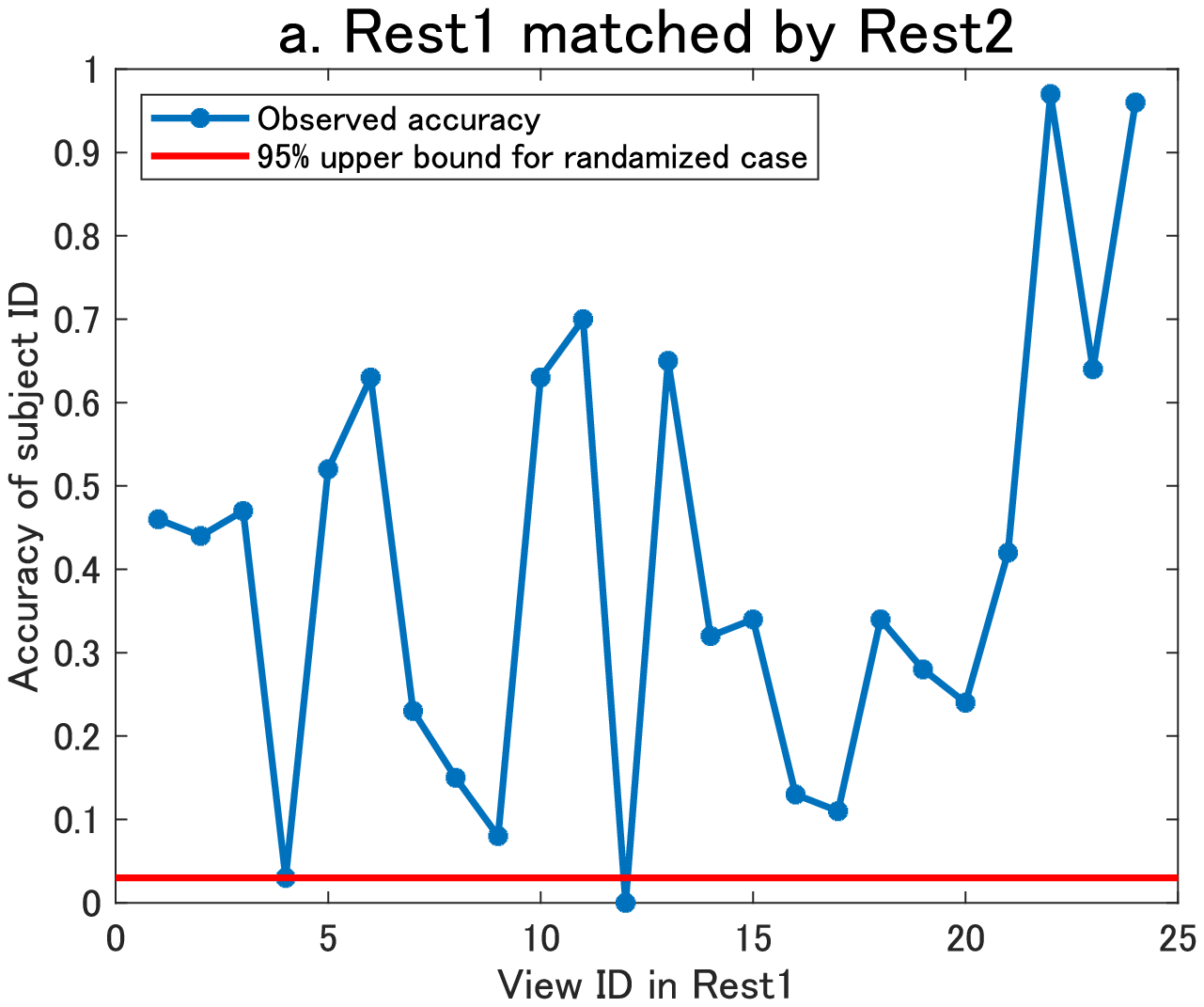}
\includegraphics[scale=0.50]{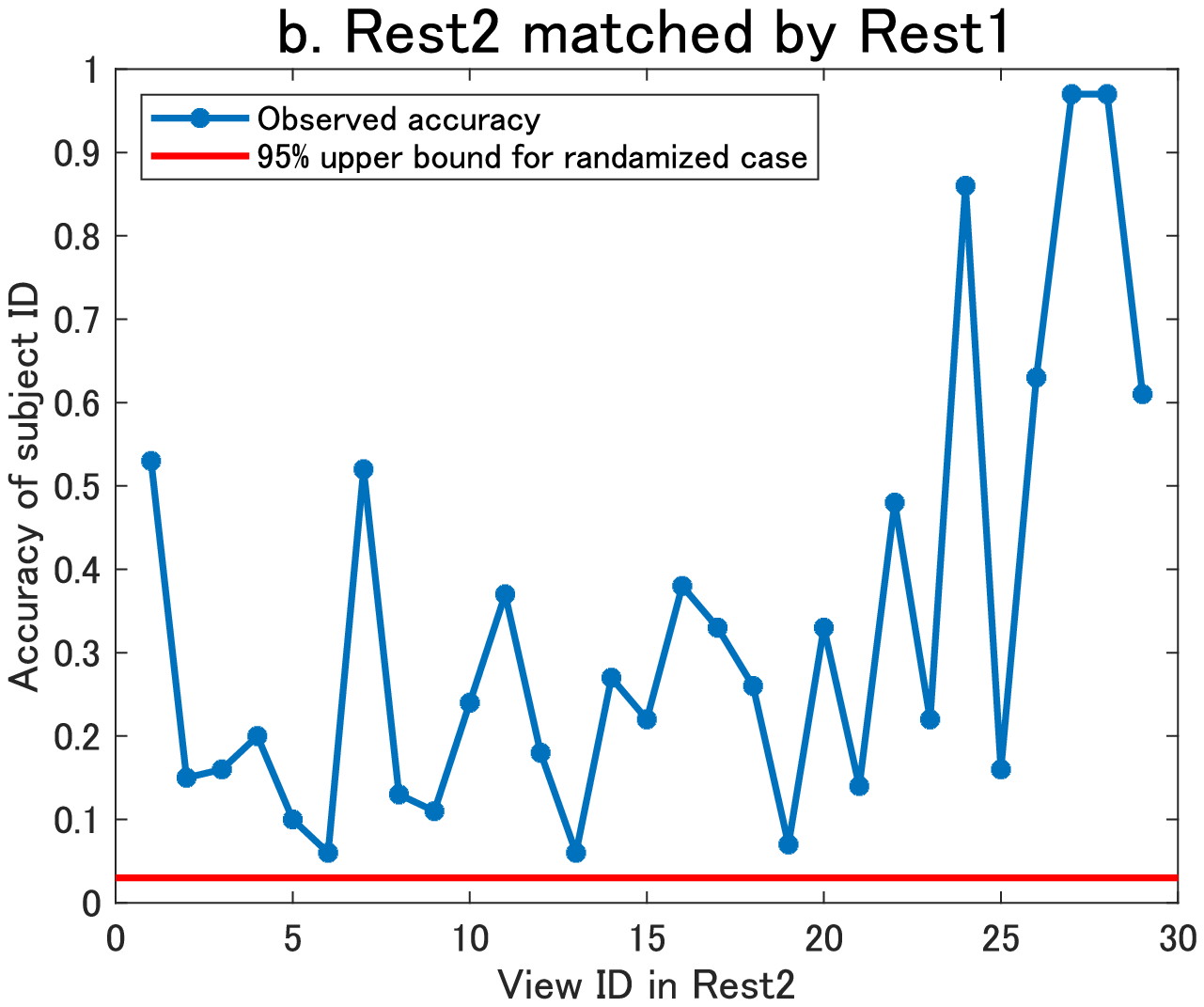}
\end{center}
\caption{Accuracy of matching of subjects between Rest1 and Rest2. Panel a: 
For a particular subject in Rest1, the \mysingleq{closest} subject in Rest2 is identified by Wishart distance, focusing on relevant ROIs for a view in Rest1. The accuracy is based on how many subjects in Rest1 are correctly matched by subjects in Rest2. 
Panel b: For a particular subject in Rest2, the \mysingleq{closest} subject in Rest1 is identified by Wishart distance, focusing on relevant ROIs for a view in Rest2. The accuracy is based on how many subjects in Rest2 are correctly matched by subjects in Rest1. 
}
\label{comparisonrestAccuracy}
\end{figure}

\begin{figure}[!]
\begin{center}
\includegraphics[scale=0.3, trim=100 100 0 80]{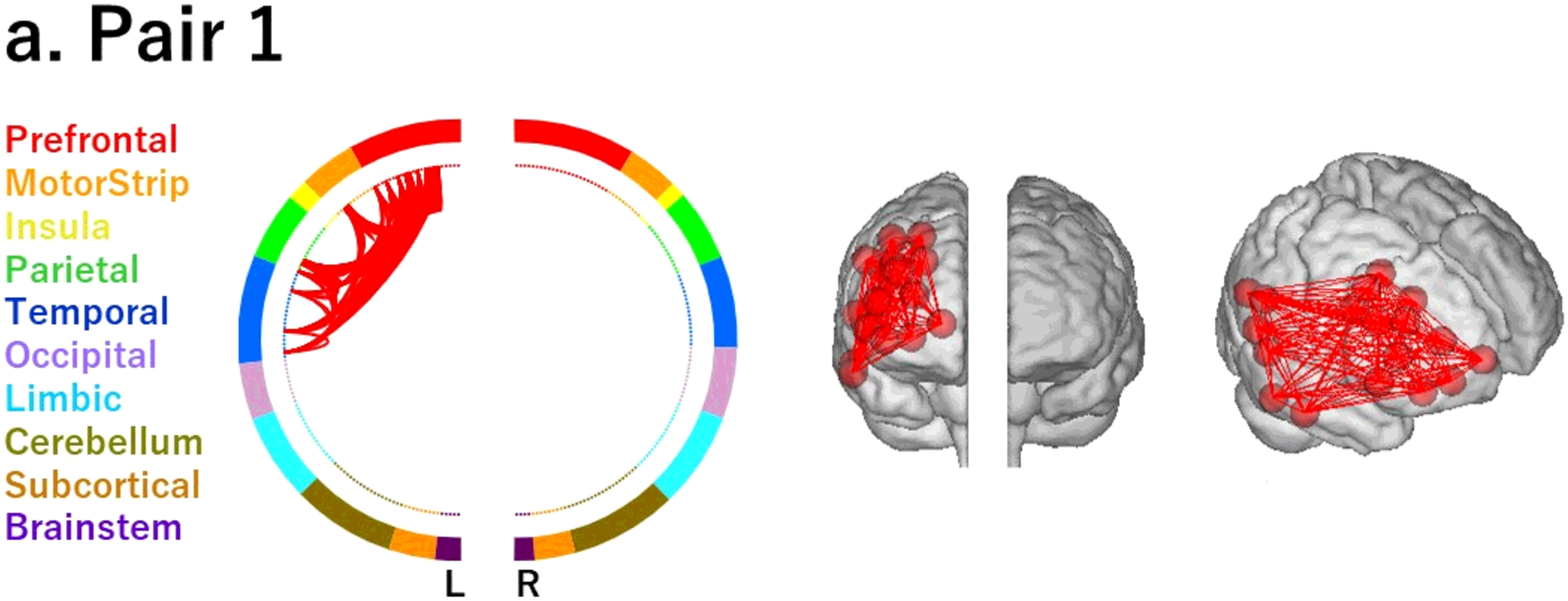}
\includegraphics[scale=0.3, trim=100 100 0 80]{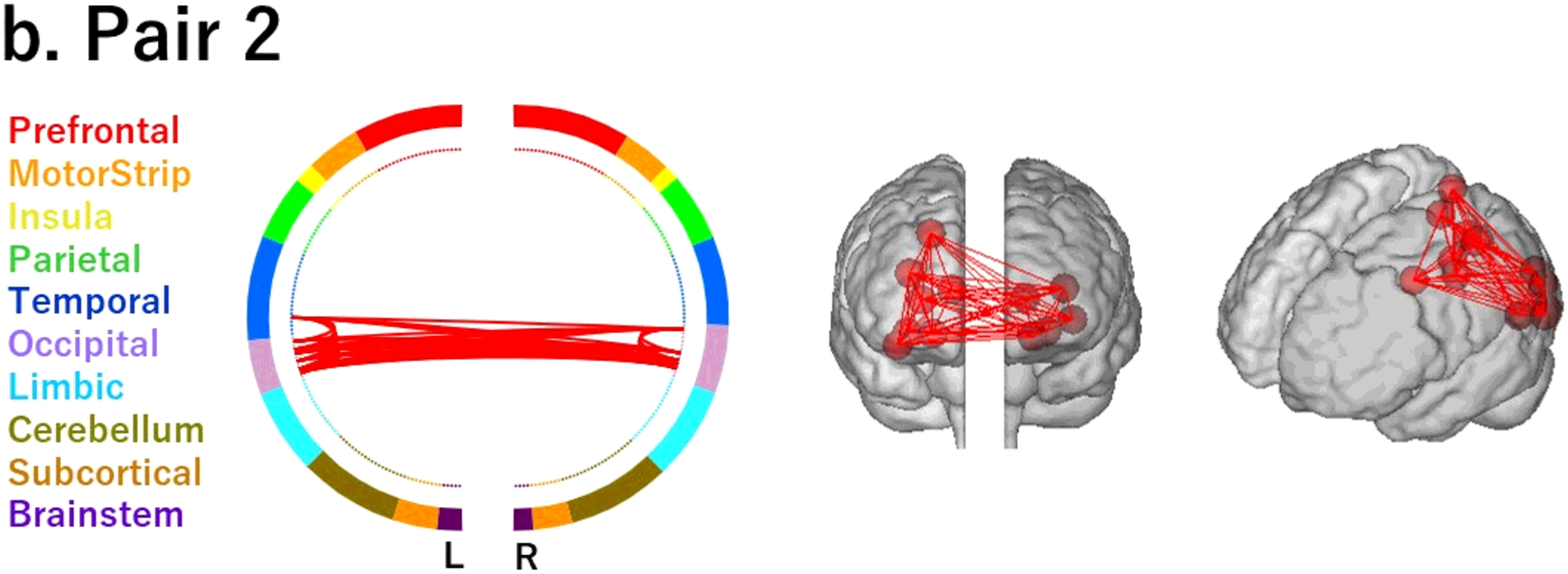}
\includegraphics[scale=0.3, trim=100 100 0 80]{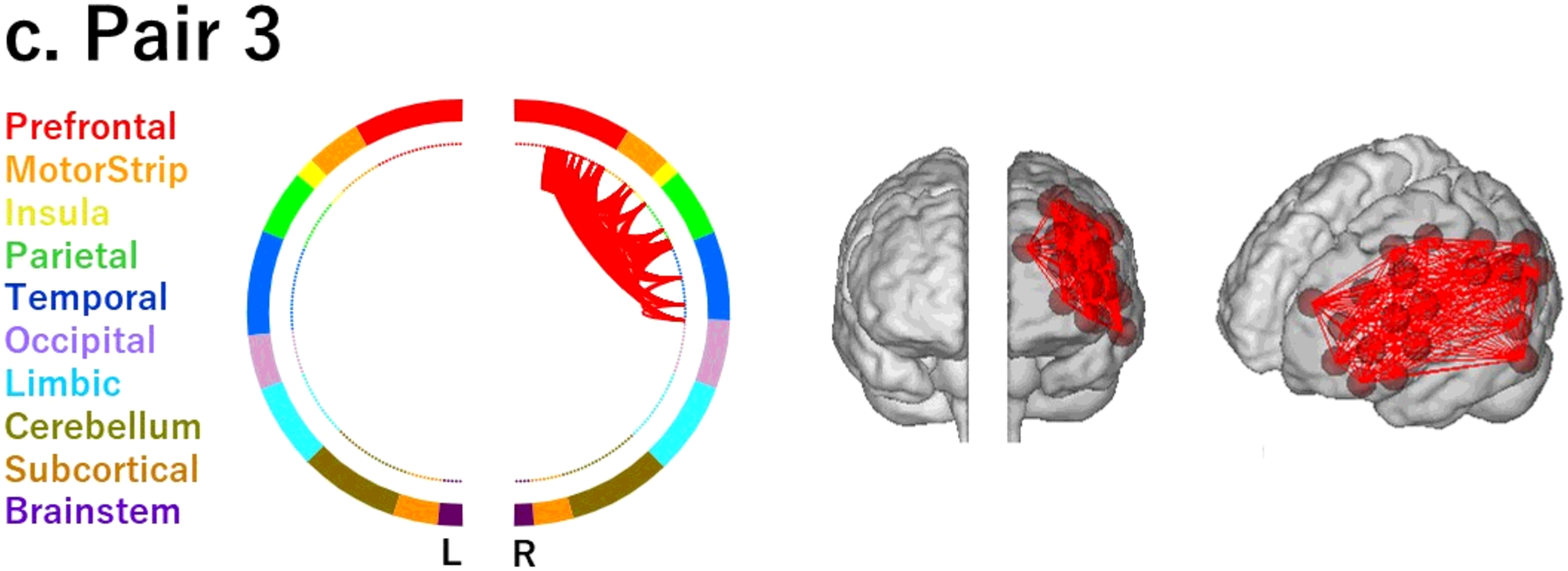}
\end{center}
\vspace{-15mm}
\caption{Relevant brain regions for three pairs of views in good agreement between Rest1 and Rest2. For pair 1 (Panel a), ROIs in view 24 of Rest1 and in view 27 of Rest2 are combined; For pair 2 (Panel b), ROIs in view 23 of Rest1 and in view 29 of Rest2; For pair 3 (Panel c), ROIs in view 22 of Rest1 and in view 28 of Rest2. Note that we used the default setting of view image in which the subject faces the screen (hence right is on the left).}
\label{comparisonrestBrain}
\end{figure}

\begin{figure}[!]
\begin{center}
\includegraphics[scale=1]{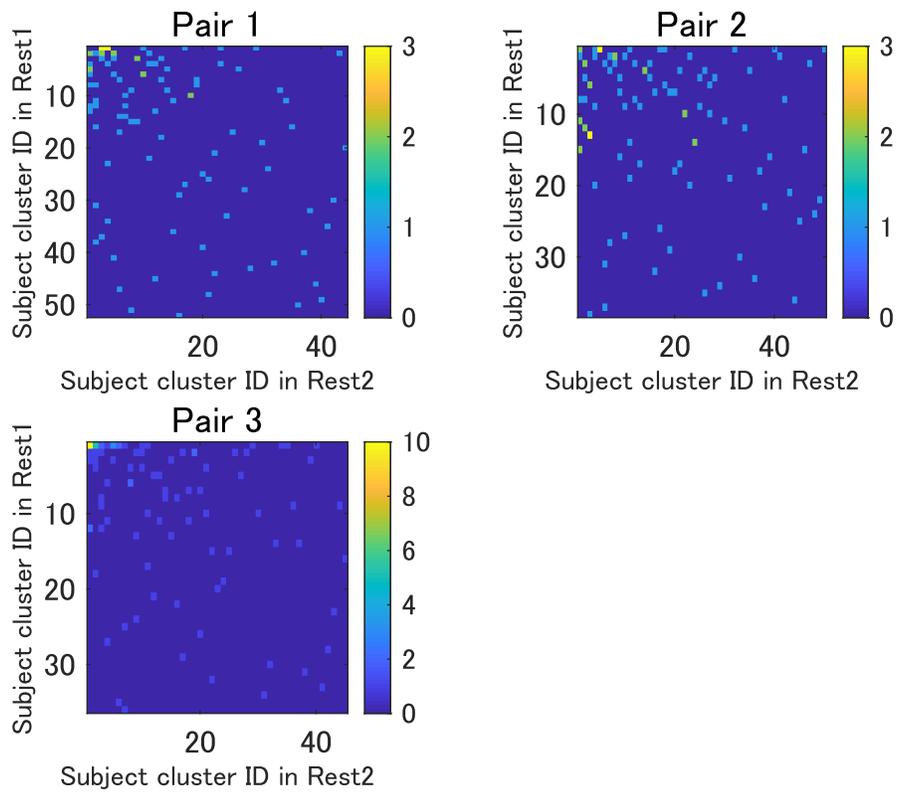}
\end{center}
\caption{Overlapping of subjects in subject clusters between Rest1 and Rest2
for pair 1, pair 2 and pair 3, respectively. The number of overlapping is denoted by color in a heatmap.}
\label{comparisonrestcrosstable}
\end{figure}

\clearpage
\begin{appendices}

\section{Node clustering within a view} \label{nodeclustering}
We assume that for each view, nodes are partitioned into several independent clusters as a view structure in Eq.(\ref{matrixstr}). 
Denoting the cluster structure of nodes in view $v$ as $\bb{y}_v$, the non-constant term in Eq.(\ref{multiden2}) becomes
\begin{eqnarray}
\no g(\bb{M}_v, T, \bb{\Sigma}_{v, k})& =& \prod_{g=1}^{G_v}
g(\bb{M}_{v, g}, T, \bb{\Sigma}_{v, k, g}),
\end{eqnarray}
where $G_v$ is the number of node clusters in view $v$; $\bb{M}_{v, g}$ covariance matrix in node cluster 
$g$ in view $v$.
Incorporating this node clustering into the model, the corresponding likelihood to Eq.(\ref{likelihood}) becomes
\begin{eqnarray}
\no && f(\bb{D}|T, \{ \bb{\Sigma}_{v, k, g} \}, \bb{u}, \{\bb{y}_v\}, \{\bb{z}_v\}) \\
\no &&~~~~~~~~~~~= \prod_{i=1}^n C_{\bb{M}^{(i)}, T} \prod_{v=1}^V \prod_{k=1}^{K_v} \prod_{g=1}^{G_v}
g(\bb{M}_{v, g}^{(i)}, T, \bb{\Sigma}_{v, k, g})^{z_v(k)}.
\label{likelihoodFull}
\end{eqnarray}

\noindent We assume a prior distribution for $\bb{y}_v$ based on CRP,
\begin{eqnarray}
\no \bb{y}_v \sim \mbox{CRP}(\alpha).
\end{eqnarray}

\noindent Finally, the posterior distribution in Eq.(\ref{lossfunction}) is given by
\begin{eqnarray}
\no L &\equiv& f(T, \bb{u}, \{\bb{y}_v\}, \{\bb{z}_v\} |
\bb{D}) \\ 
\no &&~~~~~~~= \int f(T, \{ \bb{\Sigma}_{v, k, g} \}, \bb{u}, \{\bb{y}_v\}, \{\bb{z}_v\} | \bb{D}) d \{ \bb{\Sigma}_{v, k, g} \} .
\label{lossfunction8}
\end{eqnarray}

\section{Chinese Restaurant Process (CRP)} \label{appenCRP}
A prior distribution of the view membership $\bb{u}$ ($p$-dimensional vector)
is given by a Chinese Restaurant Process (CRP) as follows \cite{gelman2013bayesian, gershman2012tutorial}:
\begin{eqnarray}
p(\bb{u}) = \frac{\alpha^K \prod_{k=1}^K (N_k-1)!}{\prod_{j=1}^p (j-1+\alpha)},
\label{CRP}
\end{eqnarray}
where $K$ is the number of views; $N_k$ the number of nodes that belong to view $k$; $\alpha$ hyperparameter (concentration parameter). An important feature of CRP is that there is no need to fix the number of views $K$ in advance. Theoretically, we can consider an infinite number of views ($K=\infty$), the optimal number being inferred in a data-driven manner based on the posterior distribution of $\bb{u}$. 
We set the concentration parameter $\alpha$ to one. For membership vectors $\bb{z}_v$ and $\bb{y}_v$, priors are defined in a similar manner based on CRP.

\section{Prior for degree of freedom $T$} \label{degree}
We assume a uniform categorical distribution for 
the prior of the degree of freedom $T$ in the Wishart distribution as follows:
\begin{eqnarray}
\no T \sim \mbox{Cat}(\cdot|\bb{c}, \bb{p}),
\end{eqnarray}
where $\bb{c}$ and $\bb{p}$ are sample space and event probability, respectively. In the present paper, we set that $\bb{c} = (T_1, \ldots, T_q)$ and that $p_i = 1/q$ ($i=1, \ldots, q$). These hyperparameters are given by 
$T_1 =p +5$, $T_2 = p+ 5 + \delta$, $\ldots$, $T_q = p+ 5 +
(q-1)\delta$,
where $p$ is the number of (all) nodes; $\delta=3$; $q = \max (n \in \mathbb{Z} ~| ~T_q \leq T^*)$. We assume the upper bound 
$T^* = \max \{ 2p, T^{ori} \}$, where $T^{ori}$ is the number of datapoints of the original time series $\bb{X}$.

\section{Algorithm}\label{appenalgo}
\subsection*{Initialization} 
We initialize relevant parameters as follows.
\begin{itemize}
\item Degree of freedom $T$: $T = \max \{2p, T^{ori} \}$, where $T^{ori}$ is the number of datapoints in the original time series data $\bb{X}$; $p$ is the number of nodes.
\item View membership $\bb{u}$: Randomly drawn from a prior distribution based on CRP 
in Eq.(\ref{CRP}).
\item Object cluster membership $\bb{z}_v$: Randomly drawn from a prior distribution based on CRP 
in Eq.(\ref{CRP}).
\item Node cluster membership $\bb{y}_v$: all ones (assuming a single node cluster per view). 
\end{itemize}

\subsection*{Optimization}
Starting from the aforementioned initial values of parameters, we iteratively update these parameters to find MAP estimates (Algorithm~\ref{algoiter}). We repeat this procedure for 1000 random initializations. As a default, we select the best model in terms of the posterior distribution. However, for the purpose of selecting a stable model, we use the optimization approach in the application to fMRI data as described in the main manuscript.

\begin{algorithm} 
\caption{MAP estimation} 
\label{alg1} 
\begin{algorithmic} 
\REQUIRE collection of $n$ covariance matrices of $p \times p$ size
\ENSURE For each of $J$ initializations, posterior value $L$, view membership $\bb{u}$, node cluster membership $\{ \bb{y}_v \}$ and object cluster membership $\{ \bb{z}_v \}$.
\vspace{5mm}
\STATE $J \leftarrow 1000$
\FOR{$j = 1$ to $J$}
\STATE Initialization : $\bb{u}$, $\{ \bb{y}_v \}$, $\{ \bb{z}_v\}$ and $T$
\STATE $L_{pre} \leftarrow 0$
\STATE $d \leftarrow 0$
\STATE $s \leftarrow 0$
\STATE $\mbox{MaxIter} \leftarrow 500$
\STATE $\mbox{MaxStability} \leftarrow 10$
\STATE $\epsilon \leftarrow 0.00001$
\WHILE{$d < \mbox{MaxIter}$ and $s<\mbox{MaxStability}$ }
\STATE Update (maximization of $L$) each element of $\bb{u}$ in a random order 
\STATE Update (maximization of $L$) each element of $\{ \bb{y}_v \}$ in a random order
\STATE Update (maximization of $L$) each element of $\{ \bb{z}_v \}$ in a random order
\STATE Update (maximization of $L$) $T$
\STATE Evaluate $L$
\IF{$ \log L - \log L_{pre} < \epsilon $}
\STATE $s=s+1$
\ELSE
\STATE $s=0$
\ENDIF
\STATE $L_{pre} = L$
\STATE $d=d+1$
\ENDWHILE
\STATE $L_j = L$, $\bb{u}_j$= $\bb{u}$, $\{ \bb{y}_v\}_j$ =$\{ \bb{y}_v \} $, $\{\bb{z}_v\}_j$=$\{ \bb{z}_v\}$
\ENDFOR
\end{algorithmic}
\label{algoiter}
\end{algorithm}
\end{appendices}

\end{document}